\documentclass[journal,10pt,final]{IEEEtran}

\usepackage[utf8]{inputenc}
\usepackage[T1]{fontenc}  			
\DeclareUnicodeCharacter{00A0}{ } 
\usepackage[english]{babel}

\usepackage{graphicx}
\usepackage[dvipsnames]{xcolor}
\usepackage{pgfplots}
\usepackage[font=small]{caption}
\usepackage{subcaption}
\usepackage{booktabs}
\usepackage{amsmath}
\usepackage{amssymb}
\usepackage{amsthm}
\usepackage{mathtools}
\usepackage[algoruled]{algorithm2e}

\SetCommentSty{mycommfont}
\usepackage{cite}
\usepackage{hyperref}
\usepackage{cleveref}				
\crefname{figure}{Fig.}{Figs.}
\crefname{algorithm}{Alg.}{Algs.}
\crefname{thm}{Theorem}{Theorems}

\usepackage{verbatim}
\usepackage{microtype}			 	
\usepackage{bm}
\usepackage{setspace}				

\usepackage{siunitx}
\usepackage{xparse}					

\theoremstyle{theorem}
\newtheorem{thm}{Theorem}

\theoremstyle{definition}
\newtheorem{defn}{Definition}

\newcommand{\toni}[1]{\textcolor{Purple}{\textbf{Toni}: #1}}
\newcommand{\jakub}[1]{\textcolor{Blue}{\textbf{Jakub}: #1}}

\newcommand{\changed}[1]{#1}

\usepackage{physics}				
\DeclareDocumentCommand \vb{ s m } {
	\IfBooleanTF #1 {
		\bm{\lowercase{#2}}
	}{
		\bm{\mathbf{\lowercase{#2}}}
	}
}
\DeclareDocumentCommand \mb { s m } {
	\IfBooleanTF #1 {
		\bm{\uppercase{#2}}
	}{
		\bm{\mathbf{\uppercase{#2}}}
	}
}
\DeclareDocumentCommand \I		{}{ ^{-1} }
\DeclareDocumentCommand \T		{}{ ^{\top} }
\DeclareDocumentCommand \IT		{}{ ^{-\top} }
\DeclareDocumentCommand \diag 	{ m }{ \mathrm{diag}\pqty{\, #1 \,} }
\DeclareDocumentCommand \eye 	{ }{ \mb{I} }
\DeclareDocumentCommand \vzero  { }{ \vb{0} }
\DeclareDocumentCommand \eVec  	{ }{ \vb{e} }

\DeclareDocumentCommand \integ 	{ O{} O{} }{ \int\limits_{#1}^{#2}\! }
\DeclareDocumentCommand \iinteg { O{} O{} }{ \iint\limits_{#1}^{#2}\! }
\DeclareDocumentCommand \iiinteg{ O{} O{} }{ \iiint\limits_{#1}^{#2}\! }
\DeclareDocumentCommand \dx 	{ O{x} }{ \,\mathrm{d}#1 }
\DeclareDocumentCommand \E 		{ O{} l m }{ \mathbb{E}_{#1}\bqty#2{ #3 } }
\DeclareDocumentCommand \V 		{ O{} l m }{ \mathbb{V}_{#1}\bqty#2{ #3 } }
\DeclareDocumentCommand \Cov 	{ O{} l m }{ \mathbb{C}_{#1}\bqty#2{ #3 } }

\DeclareDocumentCommand \SKL 	{ m m }{ \mathbb{SKL} \bqty{\left. #1 \, \middle\| \, #2 \right.} }
\DeclareDocumentCommand \N 		{ s O{\vb{x}} O{\vb{0}} O{\eye} }
{
	\IfBooleanTF{#1}{
		#2\sim\mathrm{N}\pqty{ #3,\, #4 }
	}{
		\mathrm{N}\pqty{\left. #2 \,\middle\vert\, #3,\, #4 \right.}
	}
}
\DeclareDocumentCommand \St 	{ s O{\vb{x}} O{\vb{0}} O{\eye} O{\nu} }
{
	\IfBooleanTF{#1}{
		#2\sim\mathrm{St}\pqty{#3,\, #4,\, #5}
	}{
		\mathrm{St}\pqty{\left. #2 \,\middle\vert\, #3,\, #4,\, #5 \right.}
	}
}
\DeclareDocumentCommand \G 		{ s O{s} O{\alpha} O{\beta} }
{
	\IfBooleanTF{#1}{
		#2\sim\mathrm{G}\pqty{ #3,\, #4 }
	}{
		\mathrm{G}\pqty{\left. #2 \,\middle\vert\, #3,\, #4 \right.}
	}
}
\DeclareDocumentCommand \IG 	{ s O{s} O{\alpha} O{\beta} }
{
	\IfBooleanTF{#1}{
		#2\sim\mathrm{IG}\pqty{ #3,\, #4 }
	}{
		\mathrm{IG}\pqty{\left. #2 \,\middle\vert\, #3,\, #4 \right.}
	}
}
\DeclareDocumentCommand \GP 		{ m O{\meanFcn(\vb{x})} O{\kerFcn(\vb{x}, \vb{x}')} }
{
		#1\sim\mathrm{GP}\pqty{ #2,\, #3 }
}
\DeclareDocumentCommand \pdfc	{ O{p} m m }
{
	#1\pqty{\left. #2 \,\middle\vert\, #3 \right.}
}

\DeclareDocumentCommand	\tind		{ }{ k }
\DeclareDocumentCommand	\Tind		{ }{ K }
\DeclareDocumentCommand \nlFcn 		{ s }{ \IfBooleanTF{#1}{g}{\vb{g}} }
\DeclareDocumentCommand \nlFcnU		{ s }{ \IfBooleanTF{#1}{\tilde{g}}{\tilde{\vb{g}}} }
\DeclareDocumentCommand \nlFcnT		{ s }{ \IfBooleanTF{#1}{g^\dagger}{\vb{g}^\dagger} }

\DeclareDocumentCommand \stFcn		{  }{ \vb{f} }
\DeclareDocumentCommand \msFcn		{  }{ \vb{h} }

\DeclareDocumentCommand \stVar 		{ s }{ \IfBooleanTF{#1}{x}{\vb{x}} }
\DeclareDocumentCommand \msVar		{ s }{ \IfBooleanTF{#1}{z}{\vb{z}} }
\DeclareDocumentCommand \stnVar		{ s }{ \IfBooleanTF{#1}{q}{\vb{q}} }
\DeclareDocumentCommand \msnVar		{ s }{ \IfBooleanTF{#1}{r}{\vb{r}} }
\DeclareDocumentCommand \stnCov		{ s }{ \IfBooleanTF{#1}{\sigma^2_q}{\mb{Q}} }
\DeclareDocumentCommand \msnCov		{ s }{ \IfBooleanTF{#1}{\sigma^2_r}{\mb{R}} }
\DeclareDocumentCommand \stMean		{   }{ \vb{m}^x }
\DeclareDocumentCommand \msMean		{   }{ \vb{m}^z }
\DeclareDocumentCommand \stCov		{   }{ \mb{P}^x }
\DeclareDocumentCommand \stMsCov	{   }{ \mb{P}^{xz} }
\DeclareDocumentCommand \msStCov	{   }{ \mb{P}^{zx} }
\DeclareDocumentCommand \msCov		{   }{ \mb{P}^z }
\DeclareDocumentCommand \filtMean 	{ O{\tind} }{ \stMean_{#1|#1} }
\DeclareDocumentCommand \filtCov 	{ O{\tind} }{ \stCov_{#1|#1} }
\DeclareDocumentCommand \kGain		{  }{ \mb{G} }
\DeclareDocumentCommand \inDim 		{  }{ D }
\DeclareDocumentCommand \outDim 	{  }{ E }
\DeclareDocumentCommand \idInd 		{  }{ d }
\DeclareDocumentCommand \odInd 		{  }{ e }
\DeclareDocumentCommand \stDim 		{  }{ {d_x} }
\DeclareDocumentCommand \msDim 		{  }{ {d_z} }

\DeclareDocumentCommand \inVar 		{ s }{ \IfBooleanTF{#1}{{x}}{{\vb{x}}} }
\DeclareDocumentCommand \inVarU		{ s }{ \IfBooleanTF{#1}{{\xi}}{{\vb{\xi}}} }
\DeclareDocumentCommand \inVarC		{ s }{ \IfBooleanTF{#1}{{\eta}}{{\vb{\eta}}} }
\DeclareDocumentCommand \outVar 	{ s }{ \IfBooleanTF{#1}{y}{\vb{y}} }
\DeclareDocumentCommand \inMean 	{  }{ \vb{m} }
\DeclareDocumentCommand \inCov 		{  }{ \mb{P} }
\DeclareDocumentCommand \inCovFct	{  }{ \mb{L} }

\DeclareDocumentCommand \outMean 	{  }{ \vb{\mu} }
\DeclareDocumentCommand \outCov 	{  }{ \mb{\Pi} }
\DeclareDocumentCommand \inoutCov 	{  }{ \mb{C} }
\DeclareDocumentCommand \outMeanApp	{  }{ \hat{\vb{\mu}} }
\DeclareDocumentCommand \outCovApp 	{  }{ \hat{\mb{\Pi}} }
\DeclareDocumentCommand \inoutCovApp{  }{ \hat{\mb{C}} }
\DeclareDocumentCommand \R 			{  }{ \mathbb{R} }
\DeclareDocumentCommand \Nat		{  }{ \mathbb{N} }
\DeclareDocumentCommand \D 			{  }{ \mathcal{D} }

\DeclareDocumentCommand \mtWm 		{  }{ \vb{w} }
\DeclareDocumentCommand \mtWc 		{  }{ \mb{W} }
\DeclareDocumentCommand \mtWcc 		{  }{ \mb{W}_c }
\DeclareDocumentCommand \gpExpVar	{  }{ \bar{\sigma}^2 }
\DeclareDocumentCommand \meanFcn	{  }{ m }
\DeclareDocumentCommand \kerFcn		{  }{ k }
\DeclareDocumentCommand \kerVec		{ O{\inVar} }{ \vb{\kerFcn}(#1) }
\DeclareDocumentCommand \kerMat		{  }{ \mb{K} }
\DeclareDocumentCommand \kerPar		{ s }{ \IfBooleanTF{#1}{\theta}{\vb{\theta}} }
\DeclareDocumentCommand \kerMean	{ s }{ \IfBooleanTF{#1}{\bar{k}}{\bar{\vb{k}}} }

\DeclareDocumentCommand \rbfScale   {  }{ \alpha }
\DeclareDocumentCommand \piDim 		{  }{ Q }
\DeclareDocumentCommand \piInd 		{  }{ q }
\DeclareDocumentCommand \piFcn 		{  }{ \phi }
\DeclareDocumentCommand \piCov 		{  }{ \mb{\Sigma}_{\pi} }
\DeclareDocumentCommand \vanMat		{}{ \mb{\Phi} }
\DeclareDocumentCommand \vanVec		{ O{\inVar} }{ \vb{\phi}(#1) }
\DeclareDocumentCommand \vanMean	{  }{ \bar{\vb{\phi}} }
\DeclareDocumentCommand \vanCov		{  }{ \mb{A} }
\DeclareDocumentCommand \vanCCov	{  }{ \mb{B} }
\DeclareDocumentCommand \kervanCCov	{  }{ \mb{D} }
\DeclareDocumentCommand \spNum		{  }{ N }
\DeclareDocumentCommand \spInd 		{  }{ n }
\DeclareDocumentCommand \spSet 		{  }{ \mathcal{X} }
\DeclareDocumentCommand \spMat 		{  }{ \mb{X} }
\DeclareDocumentCommand \spMatU		{  }{ \mb{\Xi} }

\DeclareDocumentCommand \fevVec 	{  }{ \vb{y} }
\DeclareDocumentCommand \fevMat		{  }{ \mb{Y} }
\DeclareDocumentCommand \degH 		{  }{ p }
\DeclareDocumentCommand \polFcn		{  }{ v }
\DeclareDocumentCommand \mulInd		{ s }{ \IfBooleanTF{#1}{\alpha}{\vb{\alpha}} }

\DeclareMathOperator{\atantwo}{\mathrm{atan2}}
\DeclareMathOperator{\spanOp}{\mathrm{span}}

\hypersetup{
	plainpages = false, 
	pdfpagelayout = OneColumn, 
	pdftitle={Improved Calibration of Numerical Integration Error in Sigma-Point Filters},
	pdfauthor={Jakub Pr\"{u}her, Toni Karvonen, Chris J.\ Oates,  Ond\v{r}ej Straka and Simo S\"{a}rkk\"{a}},
	pdfkeywords={gaussian process, bayesian quadrature, moment transformation, state estimation, nonlinear filtering},
	pdfstartview = {FitV},
	pdfdisplaydoctitle = true,
	breaklinks = true,
	pagebackref,
	linktocpage,
	bookmarks,
	bookmarksnumbered = true,
	bookmarksopen = true,
	colorlinks = true,
	linkcolor = black,
	urlcolor = black,
	citecolor = black,
	anchorcolor = black,
	hyperindex = true,
	unicode = true
}

\begin{document}

\title{Improved Calibration of Numerical Integration Error in Sigma-Point Filters}

\author{Jakub Pr\"{u}her, Toni Karvonen, Chris J.\ Oates,  Ond\v{r}ej Straka and Simo S\"{a}rkk\"{a}%
\thanks{J.\ Pr\"{u}her and O.\ Straka are with Department of Cybernetics, University of West Bohemia, Czech Republic.}%
\thanks{T.\ Karvonen and S. S\"{a}rkk\"{a} are with Department of Electrical Engineering and Automation, Aalto University, Finland.}%
\thanks{C.\ J.\ Oates is with School of Mathematics, Statistics \& Physics, Newcastle University and the Alan Turing Institute, United Kingdom.}%
}

%

\maketitle

\begin{abstract}
The sigma-point filters, such as the UKF, which exploit numerical quadrature to obtain an additional order of accuracy in the moment transformation step, are popular alternatives to the ubiquitous EKF.
The classical quadrature rules used in the sigma-point filters are motivated via polynomial approximation of the integrand, however in the applied context these assumptions cannot always be justified.
As a result, quadrature error can introduce bias into estimated moments, for which there is no compensatory mechanism in the classical sigma-point filters.
This can lead in turn to estimates and predictions that are poorly calibrated.
In this article, we investigate the Bayes--Sard quadrature method in the context of sigma-point filters, which enables uncertainty due to quadrature error to be formalised within a probabilistic model.
Our first contribution is to derive the well-known classical quadratures as special cases of the Bayes--Sard quadrature method.
\changed{Based on this, a general-purpose moment transform is developed and utilised in the design of novel sigma-point filter, which explicitly accounts for the additional uncertainty due to quadrature error.}
Numerical experiments on a challenging tracking example with misspecified initial conditions show that the additional uncertainty quantification built into our method leads to better-calibrated state estimates with improved RMSE.
\end{abstract}


\section{Introduction}\label{sec:introduction}
\IEEEPARstart{T}{his} article is concerned with quantification of uncertainty associated with sigma-point approximations, which are widely employed in nonlinear local filtering algorithms, such as the unscented Kalman filter (UKF).
The goal of filtering algorithms is to estimate the state of a dynamical stochastic system based on all measurements obtained until the present. 
The applications of filters are manifold, ranging from global positioning \cite{Grewal2007}, object tracking \cite{Haug2012,Stone2014}, simultaneous localization and mapping \cite{Durrant-Whyte2006} to weather forecasting \cite{Gillijns2006} and finance~\cite{Bhar2010}.
	
Nonlinear filtering algorithms can be categorised either as local or global. 
The global filters, such as particle filters \cite{Gustafsson2010} or point-mass filters \cite{Simandl2006}, keep track of the whole, potentially multi-modal, state posterior, which comes with increased computational demands. 
The local filters, on the other hand, have lower computational load at the cost of introducing more restrictive assumptions. 
Instead of keeping track of the whole state posterior, the local filters only work with the first two statistical moments of the state and the measurement.

For nonlinear systems and/or measurements, the moments are defined by intractable integrals that have to be approximated using numerical quadratures, also known as the sigma-point rules (which is where the filters get their name).
The classical quadrature rules, such as the Gauss--Hermite rule, are designed with the assumption that the nonlinear integrand is well-approximated with a polynomial of a given maximal degree.
Since these assumptions are almost never met in practice, there will always be a quadrature error involved.
Standard sigma-point filters do not attempt to compensate for this source of error, and in practice this can lead to estimates and predictions that are biased and over-confident \cite{Holmes2008,Lowrey2016}.

The idea to perform formal probabilistic uncertainty quantification in the numerical integration context can be traced back to \cite{Larkin1972}, with the name \emph{Bayesian quadrature} (BQ) being coined later in \cite{OHagan1991}.
The BQ has in recent years received much attention in the probabilistic numerics community \cite{Karvonen2017,Oates2016,Briol2018,Karvonen2018a}.
The BQ approach posits that the integrand can be modelled by a stochastic process defined on the domain of integration.
This model is subsequently refined by conditioning on point-wise evaluations of the integrand which induces a posterior distribution over the value of the integral.
The posterior mean of this distribution is point estimate of the value of the integral while the posterior variance expresses the integration error.

Applications of the BQ in nonlinear filtering have appeared previously in \cite{Saerkkae2016,Prueher2018} with encouraging results. 
These BQ-based filters do not generally coincide with any classical sigma-point filter, such as the UKF or Gauss--Hermite Kalman filter (GHKF), and tend to be rather sensitive to specification of the stochastic process model for the integrand.
It has been shown that classical sigma-point rules can be cast as \emph{degenerate} BQ rules~\cite{Saerkkae2016,Karvonen2017} (see~\cite{Diaconis1988} for spline methods).
This is to say that the variance associated to the integral vanishes, being thus of no use in modelling integration error.

In this article we utilise the recently proposed Bayes--Sard quadrature \cite{Karvonen2018} for the design of novel sigma-point filters, which can be viewed as probabilistic versions of the well-known sigma-point filters.
Namely, under certain conditions, the Bayes--Sard quadrature allows us to recover the classical sigma-point rules and at the same time endow the sigma-point rule with non-degenerate probabilistic output.
We thus obtain versions of standard sigma-point filters that are, to some extent, capable of accounting for numerical integration error in filtering by (in most cases) inflating the error covariance. 
In some cases, such covariance inflation is known to improve stability of nonlinear Kalman filters; see for instance~\cite[Remark 1]{Xiong2006},~\cite[Section 3.3]{Xiong2009}, and~\cite[Section V.C]{Karvonen2018b}.

The rest of the article is structured as follows. 
In \Cref{sec:sigma-point_filtering}, we formally outline the nonlinear filtering problem and the nature of sigma-point approximations.
\Cref{sec:sigma-point_moment_transforms} identifies the moment transformation problem as the central issue in local filtering and describes the structure of sigma-point moment transforms.
The Bayes--Sard quadrature is formalised in \Cref{sec:Bayes--Sard_quadrature}, which is later used in \Cref{sec:Bayes--Sard_moment_transform} to design the Bayes--Sard quadrature moment transform.
The numerical experiments are contained in \Cref{sec:numerical_experiments} and \Cref{sec:conclusion} concludes the article.

\section{Sigma-Point Filtering}\label{sec:sigma-point_filtering}
This section is devoted to the sigma-point filters, which are a subset of local filtering algorithms characterised by their reliance on a Gaussian approximation together with a numerical quadrature method.
Let the stochastic dynamical system and the process by which its state is observed be described by the state-space model
\begin{align}
	\stVar_\tind &= \stFcn(\stVar_{\tind-1}) + \stnVar_{\tind-1}, \label{eq:ssm_dynamics} \\
	\msVar_\tind &= \msFcn(\stVar_\tind) + \msnVar_\tind, \label{eq:ssm_measurement_model}
\end{align}
where the function \( \stFcn: \R^\stDim\to\R^\stDim \) is the system dynamics, \( \msFcn: \R^\stDim\to\R^\msDim \) is the measurement model, \( \stVar_\tind\in\R^\stDim \) is the latent state vector and \( \msVar_\tind\in\R^\msDim \) is the measurement vector.
Both the process noise \( \N*[\stnVar_{\tind-1}][\vzero][\stnCov] \) and the measurement noise \( \N*[\msnVar_\tind][\vzero][\msnCov] \) are zero-mean white Gaussian sequences, independent of each other and independent of the system initial condition \( \N*[\stVar_0][\stMean_0][\stCov_0] \).

The Bayesian formulation of the filtering problem can be summarized by the following two general relations.
The state posterior is
\begin{equation}\label{eq:state_posterior}
	\pdfc{\stVar_\tind}{\msVar_{1:\tind}} \propto \pdfc{\msVar_\tind}{\stVar_\tind}\pdfc{\stVar_\tind}{\msVar_{1:\tind-1}},
\end{equation}
where the likelihood \( \pdfc{\msVar_\tind}{\stVar_\tind} \) is obtained from the measurement model \labelcref{eq:ssm_measurement_model} and \( \msVar_{1:\tind} \triangleq \Bqty{\msVar_1,\, \ldots,\, \msVar_\tind} \).
The predictive density is given by the Chapman--Kolmogorov equation
\begin{equation}\label{eq:chapman-kolmogorov}
	\pdfc{\stVar_\tind}{\msVar_{1:\tind-1}} = \integ \pdfc{\stVar_\tind}{\stVar_{\tind-1}} \pdfc{\stVar_{\tind-1}}{\msVar_{1:\tind-1}} \dx[\stVar_{\tind-1}],
\end{equation}
where the transition density \( \pdfc{\stVar_\tind}{\stVar_{\tind-1}} \) is obtained from the system dynamics \labelcref{eq:ssm_dynamics}.

A vast majority of the well-known local filters, such as EKF, UKF and GHKF, can be recovered from the Bayesian formulation under a Gaussian approximation of the joint density of the state and measurement.
That is, when the density \( \pdfc{\stVar_\tind, \msVar_\tind}{\msVar_{1:\tind-1}} = \pdfc{\msVar_\tind}{\stVar_\tind}\pdfc{\stVar_\tind}{\msVar_{1:\tind-1}} \) is approximated by a Gaussian density of the form
\begin{equation}\label{eq:joint_gaussian_state_measurement}
	\N[\bmqty{\stVar_\tind \\ \msVar_\tind}][\bmqty{\stMean_{\tind|\tind-1} \\ \msMean_{\tind|\tind-1}}][\bmqty{\stCov_{\tind|\tind-1} & \stMsCov_{\tind|\tind-1}\\\msStCov_{\tind|\tind-1} & \msCov_{\tind|\tind-1}}],
\end{equation}
then the mean and covariance of the state posterior \labelcref{eq:state_posterior} have analytical form, given by
\footnote{\changed{Note that, \( \stMean_{\tind|\tind}\triangleq \E[\stVar]{\stVar_\tind \mid \msVar_{1:\tind}}\) and \( \stCov_{\tind|\tind}\triangleq \E[\stVar]{(\stVar_\tind - \stMean_{\tind|\tind})(\stVar_\tind - \stMean_{\tind|\tind})\T \mid \msVar_{1:\tind}} \)}}
\begin{align}
\filtMean &= \stMean_{\tind|\tind-1} + \kGain_\tind\pqty{\msVar_\tind - \msMean_{\tind|\tind-1}}, \label{eq:gaussian_posterior_mean} \\ 
\filtCov  &= \stCov_{\tind|\tind-1} - \kGain_\tind\msCov_{\tind|\tind-1}\kGain_\tind\T, \label{eq:gaussian_posterior_covariance}
\end{align}
where \( \kGain_\tind = \stMsCov_{\tind|\tind-1}\pqty\big{\msCov_{\tind|\tind-1}}\I \) is the Kalman gain.
The predictive moments of the state, \( \stMean_{\tind|\tind-1} \) and \( \stCov_{\tind|\tind-1} \), and the moments of measurements, \( \msMean_{\tind|\tind-1} \), \( \msCov_{\tind|\tind-1} \) and \( \stMsCov_{\tind|\tind-1} \), are defined as integrals of the form
\begin{equation}\label{eq:gaussian_integral_general_form}
	\E[\inVar]{\nlFcn(\inVar)} \triangleq \integ \nlFcn(\inVar) \N[\inVar][\inMean][\inCov] \dx[\inVar].
\end{equation}
\Cref{tab:moments} shows which quantities have to be substituted for \( \nlFcn(\stVar) \), \( \stVar \), \( \vb{m} \) and \( \mb{P} \) to obtain any of the above predictive moments.
Since the function \( \nlFcn \) being integrated is nonlinear in each case, these integrals cannot be typically computed analytically and some type of approximation needs to be employed\footnote{When the model functions \( \stFcn \) and \( \msFcn \) are linear, the integrals can be computed analytically and the \cref{eq:gaussian_posterior_mean,eq:gaussian_posterior_covariance} become identical to the Kalman filter update.}.
Each local nonlinear filter is distinguished solely by the type of integral approximation it uses.
For example, the EKF employs the first order Taylor expansion to linearise \( \nlFcn \) in the vicinity of \( \vb{m} \), which in turn facilitates analytic tractability of the moment integrals.
On the other hand, the sigma-point filters, such as the UKF and the GHKF, leverage numerical quadrature for approximation of the integral.
Since quadratures are typically designed to be used with standard Gaussian, the integrals of the form \labelcref{eq:gaussian_integral_general_form} need to be converted by employing a stochastic decoupling substitution \( \inVar^{(\spInd)} = \inMean + \inCovFct\inVarU^{(\spInd)} \), which leads to an approximation
\begin{align} 
	\E[\inVar]{\nlFcn(\inVar)} 
	&\approx \sum_{\spInd=1}^{\spNum} w_\spInd \nlFcn(\inMean + \inCovFct\inVarU^{(\spInd)}) \nonumber \\
	&= \sum_{\spInd=1}^{\spNum} w_\spInd \nlFcnU(\inVarU^{(\spInd)}), \label{eq:gaussian_integral_quadrature_decoupled}
\end{align}
\changed{where \( \inVarU^{(\spInd)} \) denotes the $ \spInd $-th unit sigma-point, $w_n \in \mathbb{R}$ is the $ \spInd $-th weight, \( \spNum \) is the total number of sigma-points, \( \inCovFct \) is a matrix factor such that \( \inCov = \inCovFct\inCovFct\T \) and \( \nlFcnU(\inVarU) \triangleq \nlFcn(\inMean + \inCovFct\inVarU) \).}
Note that various quadrature rules are distinguished by the different weights and sigma-points they prescribe to satisfy various optimality criteria.


\begin{table} 
	\centering
	\smallskip
	\begin{tabular}{ @{} >{$\,}l<{\:$} >{$\:}l<{\,$} >{$\:}l<{\,$}  >{$\:}l<{\,$} >{$\:}l<{\,$} @{} }
		\toprule
		\text{Moment}					& \nlFcn(\stVar)				& \stVar 			& \vb{m} 					& \mb{P}					\\
		\midrule
		\stMean_{\tind|\tind-1}			& \stFcn(\stVar_{\tind-1})		& \stVar_{\tind-1}	& \stMean_{\tind-1|\tind-1}	& \stCov_{\tind-1|\tind-1} 	\\
		\stCov_{\tind|\tind-1}			& \Delta\stFcn\Delta\stFcn\T	& \stVar_{\tind-1}	& \stMean_{\tind-1|\tind-1}	& \stCov_{\tind-1|\tind-1}	\\
		\msMean_{\tind|\tind-1}			& \msFcn(\stVar_\tind)			& \stVar_\tind		& \stMean_{\tind|\tind-1}	& \stCov_{\tind|\tind-1}	\\
		\msCov_{\tind|\tind-1}			& \Delta\msFcn\Delta\msFcn\T	& \stVar_{\tind}	& \stMean_{\tind|\tind-1}	& \stCov_{\tind|\tind-1}	\\
		\stMsCov_{\tind|\tind-1}		& \Delta\inVar\Delta\msFcn\T	& \stVar_{\tind}	& \stMean_{\tind|\tind-1}	& \stCov_{\tind|\tind-1}	\\
		\bottomrule
	\end{tabular}
	\caption{
		Quantities that need to be substituted into the Gaussian integral~\eqref{eq:gaussian_integral_general_form} in order to obtain every predictive moment necessary to compute the moments of the state posterior.
		The following shorthand notation is used: \( \Delta\stFcn = \stFcn(\stVar_{\tind-1}) - \stMean_{\tind|\tind-1} \), \( \Delta\msFcn = \msFcn(\stVar_{\tind}) - \msMean_{\tind|\tind-1} \), \( \Delta\inVar = \stVar_{\tind} - \stMean_{\tind|\tind-1} \).}
	\label{tab:moments}
\end{table}

\section{Sigma-Point Moment Transforms}\label{sec:sigma-point_moment_transforms}
From the above exposition, it is apparent that the central issue in local filtering is the design of the so-called moment transformations, which generate approximations to the moments of a random variable under a possibly nonlinear transformation.

Let \( \inVar\in\R^\inDim \) be an input Gaussian random variable and \( \outVar\in\R^\outDim \) an output random variable defined by
\begin{equation}\label{eq:mt_output_variable}
	\outVar = \nlFcn(\inVar) \qc \N*[\inVar][\inMean][\inCov].
\end{equation}
If the transformation \( \nlFcn \) is nonlinear, the joint density \( p(\inVar,\, \outVar) \) will be non-Gaussian in general.
However, there are many applied situations where \( \nlFcn \) is approximately linear in the region where probability mass is concentrated.
In such situations the principal error term in the moment transform is numerical quadrature error.
This error is the focus of our present work and, therefore, in what follows we proceed under the assumption that the Gaussian approximation 
\begin{equation}\label{eq:mt_joint_gaussian_input_output}
	\N[\bmqty{\inVar \\ \outVar}][\bmqty{\inMean \\ \outMean}][\bmqty{\inCov & \inoutCov \\ \inoutCov\T & \outCov}]
\end{equation}
of \( p(\inVar,\, \outVar) \) can be justified.
In this setting the moment transformation then reduces to computing the output mean \( \outMean \), covariance \( \outCov \) and cross-covariance \( \inoutCov \) as accurately as possible, when supplied with the input moments, \( \inMean \) and \( \inCov \).
This is a specific instance of \emph{uncertainty propagation}; see for example \cite{Clifford1973}.

In this article we focus on the sigma-point approximations, exemplified by \cref{eq:gaussian_integral_quadrature_decoupled}, to the moment integrals in \Cref{tab:moments}.
The well-known classical approximations, such as the Gauss--Hermite, the spherical-radial and the unscented transform, are conventionally written in the form
\begin{align}\label{eq:moments_sigma-point_classical_form}
	\outMean 	\approx	\outMeanApp 	&= \sum_{\spInd=1}^{\spNum} w_\spInd \nlFcnU(\inVarU^{(\spInd)}), \\
	\outCov 	\approx \outCovApp 		&= \sum_{\spInd=1}^{\spNum} w_\spInd \pqty\big{\nlFcnU(\inVarU^{(\spInd)}) - \outMeanApp}\pqty\big{\nlFcnU(\inVarU^{(\spInd)}) - \outMeanApp}\T, \\
	\inoutCov 	\approx	\inoutCovApp	&= \inCovFct\sum_{\spInd=1}^{\spNum} w_\spInd \inVarU^{(\spInd)} \pqty\big{\nlFcnU(\inVarU^{(\spInd)}) - \outMeanApp}\T,
\end{align}
which, under the assumption that \( \sum w_\spInd = 1\) and \( \sum w_\spInd \inVarU^{(\spInd)} = 0 \), we will prefer to write using the matrix notation as
\begin{subequations}
\begin{align} 
	\outMeanApp 	&= \fevMat\T\mtWm, \label{eq:out_mean_approx} \\
	\outCovApp 		&= \fevMat\T\mtWc\fevMat - \outMeanApp\outMeanApp\T, \label{eq:out_cov_approx} \\
	\inoutCovApp 	&= \inCovFct\spMatU\mtWcc\fevMat, \label{eq:out_ccov_approx}
\end{align}
\end{subequations}
where \( \spMatU = \bqty\big{\mqty{\inVarU^{(1)} & \ldots & \inVarU^{(\spNum)}}} \) and the matrix of integrand evaluations is given by \( \bqty{\fevMat}_{\spInd\odInd} \triangleq \nlFcnU*_\odInd(\inVarU^{(\spInd)}) \), where \( \odInd \) indexes outputs of \( \nlFcnU \) \changed{and $ \bqty{\,\cdot\,}_{\spInd\odInd} $ denotes the matrix element at position $ (\spInd, \odInd) $.} 
The vector \( \mtWm \) contains the weights and \( \mtWc = \mtWcc = \diag{\mtWm} \) for any classical sigma-point moment transform.
Each moment transform uses a different set of sigma-points and weights. 

We next discuss the unscented transform and the Gauss--Hermite quadrature in detail.

\subsection{Unscented Transform}\label{ssec:unscented_transform}

The unscented transform (UT) of \( \inDim \)-dimensional input uses \( \spNum = 2\inDim + 1 \) sigma-points, which exploit symmetry of the Gaussian distribution, given, for \( \idInd = 1, \ldots, \inDim \), by
\begin{equation}\label{eq:UT-sigma-points}
	\inVarU^{(0)} = \vzero \qc \inVarU^{(\idInd)} = \sqrt{c}\,\eVec_\idInd \qc \inVarU^{(\inDim + \idInd)} = -\sqrt{c}\,\eVec_{\inDim + \idInd},
\end{equation}
where \( \eVec_\idInd \) is the standard unit vector and \( c = \inDim + \kappa \) for a scaling parameter \( \kappa \).
The weights are defined as
\begin{equation}\label{eq:UT-weights}
	w_0 = \frac{\kappa}{c} \qc w_\idInd = \frac{1}{2c} \qc w_{\inDim + \idInd} = \frac{1}{2c}.
\end{equation}
This selection of sigma-points and weights yields a quadrature rule that integrates exactly all polynomials of (total) degree at most three; the derivation is essentially contained in the proof of \Cref{thm:ukf}.
The \emph{spherical-radial rule}, which is used in the cubature Kalman filter (CKF) \cite{Arasaratnam2009}, is equivalent to the UT with \( \kappa = 0 \); it therefore lacks the central sigma-point. 

\subsection{Gauss--Hermite Rule}\label{ssec:gauss-hermite_rule}

From non-singularity of the Vandermonde matrix $[\mb{V}]_{nm} = x_n^{m-1}$ for any distinct sigma-points $x_1, \ldots, x_p \in \mathbb{R}$ it follows that there are unique weights such that $\sum_{n=1}^p w_n x_n^{m} = \int x^m \mathrm{N}(0,1) \mathrm{d} x$ for every $m \leq p-1$ (i.e., the rule has a degree of exactness $p-1$).
However, degree of exactness $2p-1$ can be achieved with $p$ sigma-points if these are selected to be the roots of the \( \degH \)-th degree Hermite polynomial \( \mathrm{H}_\degH \). 
The weights are then given by
\begin{equation}\label{eq:gh_weights_1d}
	w_\spInd = \frac{\degH !}{\degH^2 \mathrm{H}_{\degH-1}(\inVarU*^{(\spInd)})^2}.
\end{equation}
This is the Gauss--Hermite (GH) rule \cite{Arasaratnam2007,Wu2006,Ito2000,Golub1969}.
In multivariate versions, the sigma-points are formed as Cartesian products of the aforementioned one-dimensional points and the weights are products of $w_\spInd$ in \cref{eq:gh_weights_1d}.
The multivariate GH rule exactly integrates functions in the space
\begin{equation} \label{eq:gh-poly-space}
\Pi_{2\degH-1}^\text{max} \triangleq \spanOp \Bqty\Big{ \inVar^{\vb{\alpha}} \, \colon \, \vb{\alpha} \in \Nat_0^\inDim, \, \max_{\idInd=1,\ldots,\inDim} \alpha_\idInd \leq 2\degH-1 },
\end{equation}
where \( \inVar^{\vb{\alpha}} = \prod_{\idInd=1}^{\inDim}\inVar*_\idInd^{\alpha_\idInd} \) denotes multivariate monomial. 
Because of the Cartesian product design, the number of points, \( \spNum = \degH^\inDim \), in the GH rule grows exponentially with dimension, which makes it practically unattractive for \( \inDim > 5 \) \cite{Wu2006}. 
The problem can be partially mitigated by using sparse grids~\cite{Jia2012}.

\section{Bayesian Quadrature}\label{sec:Bayes--Sard_quadrature}
This section reviews the underlying philosophy of the Bayesian quadrature as an alternative perspective on numerical integration and describes the Bayes--Sard quadrature as a necessary stepping stone on the way to building the Bayes--Sard moment transform proposed in \Cref{sec:Bayes--Sard_moment_transform}.
A general formulation of the BQ is presented for integrals 
\begin{equation}\label{}
	\E[\inVar]{\nlFcnT*(\inVar)} = \integ \nlFcnT*(\inVar)p(\inVar) \dx[\inVar]
\end{equation}
with arbitrary density function $p$. 
Vector-valued integrands are discussed in \Cref{ssec:bsq_vector-valued_integrands}.
The moment transform proposed in \Cref{sec:Bayes--Sard_moment_transform} then specialises to the case \( p(\inVar) = \N \).
Throughout this section, the true integrand will be denoted by \( \nlFcnT* \) to distinguish it from the stochastic model of the integrand.

From \cref{eq:gaussian_integral_quadrature_decoupled} it is clear that the quadrature approximation of the integral \labelcref{eq:gaussian_integral_general_form} is based on limited knowledge about the behaviour of the integrand, because it only relies on finitely many evaluations.
The design of classical quadrature rules typically involves formation of polynomial interpolant passing through the observed function values, which is then integrated instead of the intractable integrand.
\changed{
	The polynomial interpolation of the integrand consequently implies that the classical rules are only able to integrate polynomial integrands exactly. 
	Another downside of the classical rules is that they are unable to account for the functional uncertainty (interpolation error), which occurs when the integrand is not a polynomial.
}

The Bayesian approach to quadrature \cite{Larkin1972,OHagan1991,Briol2018} (see \cite{Larkin1970,SommarivaVianello2006,Oettershagen2017} for equivalent non-probabilistic formulations) aims to address these limitations by treating the numerical approximation of intractable integrals as a problem of Bayesian statistical inference, where a prior for the integrand is specified by a stochastic process model \( \nlFcn*(\inVar) \) with user-defined mean function \( \meanFcn(\inVar) = \E[\nlFcn*]{\nlFcn*(\inVar)} \) and covariance (or kernel) function \( \kerFcn(\inVar, \inVar') = \Cov[\nlFcn*]{\nlFcn*(\inVar), \nlFcn*(\inVar')} \)\changed{, where $ \inVar' $ denotes the second argument (not a transpose).}
The dataset \( \D = \Bqty{(\inVar^{(\spInd)}, \nlFcnT*(\inVar^{(\spInd)}))}_{\spInd=1}^\spNum \) comprises evaluations of the integrand \( \nlFcnT*(\inVar^{(\spInd)}) \) at pre-defined points \( \inVar^{(\spInd)} \).
Conditioning on \( \D \) leads to a posterior stochastic process, with mean \( \meanFcn_\D(\inVar) = \E[\nlFcn*\mid\D]{\nlFcn*(\inVar)} \) and covariance \( \kerFcn_\D(\inVar, \inVar') = \Cov[\nlFcn* \mid \D]{\nlFcn*(\inVar), \nlFcn*(\inVar')} \),  which in turn induces a posterior marginal distribution on the value of the integral \( \E[\inVar]{\nlFcn*(\inVar)} \), with the first two moments given by \cite{Rasmussen2003}
\begin{align}
	\E[\nlFcn* \mid \D]{\E[\inVar]{\nlFcn*(\inVar)}} &= \E[\inVar]{\E[\nlFcn* \mid \D]{\nlFcn*(\inVar)}}, \label{eq:bq_integral_mean} \\
	\V[\nlFcn* \mid \D]{\E[\inVar]{\nlFcn*(\inVar)}} &= \E[\inVar,\inVar']{\Cov[\nlFcn* \mid \D]{\nlFcn*(\inVar), \nlFcn*(\inVar')}}. \label{eq:bq_integral_variance}
\end{align}
The mean is a convenient point estimate while the full posterior serves as a probabilistic model of the integration error.
The most common stochastic process model of the integrand is a Gaussian process (GP), which has been studied extensively \cite{Briol2018,Rasmussen2006}.
The standard formulation of the GP regression model, which is limited to modelling of scalar functions \( \nlFcnT*:\R^\inDim\to\R \), has been extended to vector functions in several ways \cite{Deisenroth2009,Alvarez2012,Xi2018}.
Alternatively, one could use the \( t \)-process (TP) regression \cite{Shah2014,Prueher2017a}, which offers more flexible uncertainty modelling capabilities and contains the GP as a special case.

In the next section, we specify the Bayes--Sard GP model which is later used to construct the Bayes--Sard quadrature in \Cref{ssec:bayes-sard_quadrature}.


\subsection{Bayes--Sard Gaussian Process Model}\label{ssec:bayes-sard_gp_model}

Let $\pi$ be a linear function space spanned by $\piDim \leq \spNum$ functions $\piFcn_1,\, \ldots,\, \piFcn_\piDim \colon \R^\inDim \to \R$. 
Modeling of the scalar integrand \( g^\dagger:\R^\inDim\to\R \) in Bayes--Sard quadrature (BSQ) begins by considering a hierarchical GP prior given by
\begin{align}
	\vb{\gamma} 		&\sim \mathrm{N}(\vzero, \piCov), \label{eq:bsq_hierarchical_bsgp_prior_start} \\
	\meanFcn(\inVar) 	&= \sum_{\piInd=1}^{\piDim} \gamma_\piInd \piFcn_\piInd(\inVar), \\
	\nlFcn*(\inVar) 	&\sim \mathrm{GP}(\meanFcn(\inVar),\, \kerFcn(\inVar, \inVar'; \kerPar)), \label{eq:bsq_hierarchical_bsgp_prior_end}
\end{align}
where the prior mean function \( \meanFcn(\inVar): \R^\inDim\to\R \) is composed of basis functions \( \piFcn_\piInd(\inVar) \) of \( \piDim \)-dimensional linear space \( \pi \) and the prior covariance function (kernel) \( \kerFcn(\inVar, \inVar'; \kerPar): \R^\inDim\times\R^\inDim\to\R \) can be any symmetric positive-definite function parametrized by the vector \( \kerPar \) \changed{(see \cref{ssec:bsq_kernel_choice} for concrete example)}.
The dependence on \( \kerPar \) will be tacitly assumed and explicitly denoted only when required.
Discussion about the particular choice of the kernel and its effects is postponed to \Cref{ssec:bsq_kernel_choice}.
The above model differs from the one often used in Gaussian process based Bayesian quadrature in that the prior mean function is non-zero and, furthermore, its coefficients are random. 

The next phase in modelling is to consider a flat prior limit on the mean function coefficients, such that \( \piCov\to\infty \)~\cite[Chapter 4]{Santner2003}.
\changed{
	In order for the GP posterior to be well-defined (i.e., to guarantee the inverse matrices in~\crefrange{eq:bsgp_posterior_mean1}{eq:bsgp_posterior_covariance} exist), the set \( \spSet = \{\inVar^{(1)},\, \ldots,\, \inVar^{(\spNum)} \}\) of sigma-points must meet the following condition of \( \pi \)-unisolvency, which is related to existence of interpolants formed out of linear combinations of $\phi_1, \ldots, \phi_\piInd$. 
	For instance, for any $N$ distinct points in one dimension and any function defined on these points it is possible to construct a unique polynomial interpolant of degree at most $N-1$. 
	Unfortunately, in higher dimensions there are point sets for which this does not hold; hence the need for a unisolvency condition. See~\cite[Supplement~B]{Karvonen2018} for more details.
}
\begin{defn}[\( \pi \)-unisolvency]
	Let \( \pi \) be a \( \piDim \)-dimensional linear space spanned by \( \Bqty{\phi_1, \ldots, \phi_\piInd} \).
	A point set \( \spSet \) is said to be \( \pi \)-unisolvent if and only if the \( \spNum\times\piDim \) alternant matrix \( \bqty{\vanMat}_{\piInd\spInd} \triangleq \phi_\piInd(\inVar^{(\spInd)}) \) is of full-rank.
\end{defn}

Following the Bayesian paradigm, the final step is to condition the GP on the set of sigma-points \( \spSet \) and the corresponding integrand evaluations, collectively denoted as \( \D = \Bqty{(\inVar^{(\spInd)}, \nlFcn*^\dagger(\inVar^{(\spInd)}))}_{\spInd=1}^\spNum \), to arrive at the GP posterior.
The GP posterior in the flat prior limit, for \( \pi \)-unisolvent \( \spSet \), becomes~\cite{Karvonen2018}
\begin{align}
	\E[\nlFcn*\mid\D]{\nlFcn*(\inVar)} 
	&= \kerVec[\inVar]\T\kerMat\I\fevVec \nonumber \\
	&\mathrel{\phantom{=}} - \vb{\psi}(\inVar)\T\bqty\big{\vanMat\T\kerMat\I\vanMat}\I\vanMat\T\kerMat\I\fevVec, \\
	\Cov[\nlFcn*\mid\D]{\nlFcn*(\inVar), \nlFcn*(\inVar')} \label{eq:bsgp_posterior_mean1}
	&= \kerFcn(\inVar, \inVar') - \kerVec[\inVar]\T\kerMat\I\kerVec[\inVar'] \nonumber \\
	&\mathrel{\phantom{=}} + \vb{\psi}(\inVar)\T\bqty\big{\vanMat\T\kerMat\I\vanMat}\I\vb{\psi}(\inVar'),
\end{align}
where \( \vb{\psi}(\inVar) = \kerVec[\inVar]\kerMat\I\vanMat - \vanVec[\inVar] \), \( \bqty{\kerVec}_\spInd \triangleq \kerFcn(\inVar, \inVar^{(\spInd)}) \), \( \bqty{\vanVec}_\piInd \triangleq \piFcn_\piInd(\inVar) \) and \( \bqty{\fevVec}_\spInd \triangleq \nlFcn*^\dagger(\inVar^{(\spInd)}) \)\changed{, where $ \bqty{\,\cdot\,}_n $ denotes the $ n $-th element of the given vector.}
It is worth pointing out that all sigma-point sets in the established classical filters are \( \pi \)-unisolvent. 

We further restrict the model to the case when \( \spNum = \piDim \), which means the alternant matrix \( \vanMat \) is square and, due to $\pi$-unisolvency of $\spSet$, invertible.
This leads to the final form of the posterior mean and covariance of the Bayes--Sard GP model:
\begin{align}
	\E[\nlFcn*\mid\D]{\nlFcn*(\inVar)} 
	&= \vanVec[\inVar]\T\vanMat\I\fevVec, \label{eq:bsgp_posterior_mean} \\
	\Cov[\nlFcn*\mid\D]{\nlFcn*(\inVar), \nlFcn*(\inVar')}
	&= \kerFcn(\inVar, \inVar') - 2\kerVec[\inVar]\T\vanMat\IT\vanVec[\inVar'] \nonumber \\
	&\mathrel{\phantom{=}} + \vanVec[\inVar]\T\bqty\big{\vanMat\T\kerMat\I\vanMat}\I\vanVec[\inVar']. \label{eq:bsgp_posterior_covariance}
\end{align}
Note that the posterior mean now only depends on the choice of the function space \( \pi \) and the kernel affects only the posterior covariance.

\subsection{Vector-Valued Integrands}\label{ssec:bsq_vector-valued_integrands}
Until now, we have only considered scalar-valued integrands. 
The model specified by \cref{eq:bsgp_posterior_mean,eq:bsgp_posterior_covariance} can be straightforwardly extended to vector-valued integrands \( \nlFcn^\dagger:\R^\inDim\to\R^\outDim \) that comply with the specification of the moment transformation problem in \cref{eq:mt_output_variable}.
Noticing that we can decompose the integrand as \( \nlFcn^\dagger(\inVar) = \bmqty{\nlFcn*^\dagger_1(\inVar) & \ldots & \nlFcn*^\dagger_\outDim(\inVar)}\T \), the simplest solution is to use \cref{eq:bsgp_posterior_mean,eq:bsgp_posterior_covariance} to model each coordinate function independently, either using a common kernel parameter for all outputs, which is accomplished by
\begin{equation}\label{eq:bsgp_multi-output_single_parameter_posterior}
	 \GP{\nlFcn*_\odInd(\inVar)\mid\D}[\meanFcn_\D(\inVar)][\kerFcn_\D(\inVar, \inVar'; \kerPar)],
\end{equation}
or using a different kernel parameter values for each output, so that
\begin{equation}\label{eq:bsgp_multi-output_multi_parameter_posterior}
	\GP{\nlFcn*_\odInd(\inVar)\mid\D}[\meanFcn_\D(\inVar)][\kerFcn_\D(\inVar, \inVar'; \kerPar_\odInd)]
\end{equation}
for all \( \odInd = 1,\, \ldots,\, \outDim \).
In both cases, the GP posterior mean function is given as
\begin{equation}\label{eq:bsgp_multi-output_posterior_mean}
	\vb{\meanFcn}_\D(\inVar) \triangleq \E[\nlFcn\mid\D]{\nlFcn(\inVar)} = \fevMat\T\vanMat\I\vanVec[\inVar],
\end{equation}
where \( \bqty{\fevMat}_{\spInd\odInd} = \nlFcn*^\dagger_\odInd(\inVar^{(\spInd)}) \).
For the single-parameter model \labelcref{eq:bsgp_multi-output_single_parameter_posterior} the posterior covariance becomes
\begin{equation}\label{eq:bsgp_multi-output_single_parameter_posterior_covariance}
	\kerMat_\D(\inVar, \inVar') \triangleq \Cov[\nlFcn\mid\D]{\nlFcn(\inVar), \nlFcn(\inVar')} = \kerFcn_\D(\inVar, \inVar'; \kerPar)\, \eye_\outDim
\end{equation}
and for the multi-parameter model \labelcref{eq:bsgp_multi-output_multi_parameter_posterior}, we get
\begin{equation}\label{eq:bsgp_multi-output_multi_parameter_posterior_covariance}
	\vphantom{\Big(} \kerMat_\D(\inVar, \inVar') = \diag{\bqty\big{\mqty{\kerFcn_\D(\inVar, \inVar'; \kerPar_1) & \ldots & \kerFcn_\D(\inVar, \inVar'; \kerPar_\outDim)}}}.
\end{equation}
Both of these modelling choices assume that the outputs are conditionally independent given the inputs.
Alternatively, the use of multi-output Gaussian processes~\cite{Alvarez2012,Xi2018} would make it possible to model correlations between coordinate functions and use coordinate-dependent sigma-points at the expense of increased computational cost.

\subsection{Bayes--Sard Quadrature}\label{ssec:bayes-sard_quadrature}


The advantage of using a GP for modelling the integrand is that as it gets transformed by the integral, which is a linear operator, the resulting distribution over the value of the integral is also Gaussian\footnote{Analogous invariance result holds for the TP model as well.}.
The Bayes--Sard quadrature~\cite{Karvonen2018,Larkin1974,OHagan1991} enables enforcing exactness conditions of the form 
\begin{equation*}
	\E[\nlFcn\mid\D]{\E[\inVar]{\nlFcn(\inVar)}} = \integ \nlFcnT(\inVar) p(\inVar) \dx[\inVar]
\end{equation*}
for all functions $\nlFcnT \colon \mathbb{R}^D \to \mathbb{R}^E$ such that $\nlFcnT*_e \in \pi$ for each $e=1,\ldots,E$.
As shown in \Cref{ssec:bsq_relationship_classical_rules}, the classical quadrature methods can be replicated by judicious choice of the function space \( \pi \).
The posterior integral mean and variance under the Bayes--Sard quadrature are straightforwardly derived by plugging the Bayes--Sard GP model moments from \cref{eq:bsgp_multi-output_posterior_mean,eq:bsgp_multi-output_multi_parameter_posterior_covariance} into the general BQ expressions in \cref{eq:bq_integral_mean,eq:bq_integral_variance}.

For the mean of the posterior distribution of the integral, we have
\begin{align}
	\E[\nlFcn\mid\D]{\E[\inVar]{\nlFcn(\inVar)}} = \E[\inVar]{\vb{\meanFcn}_\D(\inVar)}	= \fevMat\T\vanMat\I\vanMean, \label{eq:bsq_integral_mean}
\end{align}
where 
\begin{equation}
\bqty{\vanMean}_\piInd = \E[\inVar]{\piFcn_\piInd(\inVar)} = \integ \piFcn_\piInd(\inVar) p(\inVar) \dx[\inVar] .
\end{equation}
Recognizing that the vector of quadrature weights is \( \mtWm = \vanMat\I\vanMean \), we see that the posterior mean of the integral
\begin{equation}\label{eq:bsq_integral_mean_as_weighted_sum}
	\E[\nlFcn\mid\D]{\E[\inVar]{\nlFcn(\inVar)}} = \fevMat\T\mtWm = \sum_{\spInd=1}^{\spNum} w_\spInd \nlFcn^\dagger(\inVar^{(\spInd)}),
\end{equation}
takes on the form of weighted sum from \cref{eq:gaussian_integral_quadrature_decoupled}.
The integral covariance becomes
\begin{equation}
	\vphantom{ \Bigg( } 
	\V[\nlFcn\mid\D]{\E[\inVar]{\nlFcn(\inVar)}} 
	= \E[\inVar, \inVar']{\kerMat_\D(\inVar, \inVar')} = \diag{\bmqty{\kerMean*^1_\D & \ldots & \kerMean*^\outDim_\D}} \label{eq:bsq_integral_covariance_general}
\end{equation}
where
\begin{align}
	\kerMean*^e_\D 
	&\triangleq \E[\inVar, \inVar']{\kerFcn_\D(\inVar, \inVar'; \kerPar_\odInd)} \nonumber \\
	&= \kerMean* - 2\kerMean\T\vanMat\IT\vanMean  + \vanMean\T\bqty\big{\vanMat\T\kerMat\I\vanMat}\I\vanMean. \label{eq:bsq_integral_covariance}
\end{align}
Since the single-parameter model in \cref{eq:bsgp_multi-output_single_parameter_posterior_covariance} is a special case of \cref{eq:bsgp_multi-output_multi_parameter_posterior_covariance}, the posterior integral variance under this model would be a trivial modification of \cref{eq:bsq_integral_covariance_general}.

\subsection{Relationship to Classical Sigma-Point Rules}\label{ssec:bsq_relationship_classical_rules}

As stated in the previous section, careful selection of $\pi$ (via the basis functions \( \piFcn_q \)) allows for recovery of many well-known classical quadrature rules used in nonlinear filtering.
Below, we show that the unscented transform and the Gauss--Hermite rule are special cases of the BSQ whenever the space \( \pi \) is spanned by suitably selected polynomial basis.
Similar results can be proved for many other sigma-point rules.
Note that the BSQ reports a non-zero integral variance even for \( \nlFcn^\dagger \) whose coordinate functions are in $\pi$ (and hence integrated exactly).
This behaviour is desirable because, given only a finite set of function values, one can never tell with certainty the true nature of the integrand.

\begin{thm}\label{thm:ukf}
	Consider the standard Gaussian distribution, \( p(\inVar) = \N \).
	Select the $2\inDim + 1$ dimensional function space
	\begin{equation}
		\pi = \spanOp \Bqty{1,\, x_1,\, \ldots,\, x_\inDim,\, x_1^2,\, \ldots,\, x_\inDim^2}
	\end{equation}
	and the $ \spNum = 2\inDim + 1$ unscented transform points~\labelcref{eq:UT-sigma-points}.
	Then, the Bayes--Sard weights $\mtWm = \vanMat\I\vanMean $ that determine the posterior mean~\labelcref{eq:bsq_integral_mean} coincide with the unscented transform weights~\labelcref{eq:UT-weights}.
\end{thm}
\begin{IEEEproof} 
	Because $\dim(\pi) = \spNum$, the Bayes--Sard weights \( \mtWm \) solve the linear system $\vanMat\mtWm = \vanMean$. That is, they are the unique weights such that
	\begin{equation}\label{eq:UT-condition}
		\sum_{\spInd=0}^{2\inDim} w_\spInd \polFcn(\inVarU^{(\spInd)}) = \integ \polFcn(\inVar) \N \dx[\inVar]
	\end{equation}
	for every polynomial $\polFcn \in \pi$.
	In the following, let \( \idInd = 1,\, \ldots,\, \inDim \).
	We have $\integ \N \dx[\inVar] = 1$ and
	\begin{equation}
		\integ x_\idInd\, \N \dx[\inVar] = 0, \quad \integ x_\idInd^2\, \N \dx[\inVar] = 1.
	\end{equation}
	Consequently,~\cref{eq:UT-condition} is equivalent to
	\begin{equation}
		\sum_{\spInd=0}^{2\inDim} w_{\spInd} = 1, \quad \sum_{\spInd=0}^{2\inDim} w_{\spInd} x_{\spInd,\idInd} = 0, \quad \sum_{\spInd=0}^{2\inDim} w_{\spInd} x_{\spInd,\idInd}^2 = 1
	\end{equation}
	Because $\vb{\xi}_\idInd = -\vb{\xi}_{\inDim + \idInd}$, the second of these equations implies that $w_{d} = w_{D + d}$, while the third one yields $w_{\idInd} = w_{\inDim + \idInd} = 1/(2c)$.
	Furthermore, $w_0 = \kappa/c$ due to the weights summing up to one.
	We have thus solved the BSQ weights $\mtWm = \vanMat\I\vanMean$ and see that they are precisely UT weights in~\cref{eq:UT-weights}.
\end{IEEEproof}

\begin{thm}\label{thm:gh}
	Consider the standard Gaussian distribution, $ p(\inVar) = \N $, and let $ \degH \geq 1 $.
	Select the $ \degH^\inDim $ dimensional function space \changed{$ \pi = \Pi_{\degH-1}^\text{max} \triangleq \spanOp \Bqty{ \inVar^{\vb{\alpha}} \, \colon \, \vb{\alpha} \in \Nat_0^\inDim, \, \max_{\idInd=1,\ldots,\inDim} \alpha_\idInd \leq \degH-1 } $}, and the points that constitute the Cartesian product of the roots of the \( \degH \)-th degree Hermite polynomial.
	Then, the Bayes--Sard weights $ \mtWm = \vanMat\I\vanMean $ that determine the posterior mean~\labelcref{eq:bsq_integral_mean} coincide with the classical Gauss--Hermite weights from \Cref{ssec:gauss-hermite_rule}.
\end{thm}
\begin{IEEEproof}
	Since the Bayes--Sard weights yield, by their definition, a quadrature rule exact for functions in $\pi$ and it is known that, given the Gauss--Hermite points, the Gauss--Hermite weights are the unique weights that determine a quadrature rule exact for this very same function space (see \Cref{ssec:gauss-hermite_rule}), the result follows.
\end{IEEEproof}

\subsection{Choice of Kernel}\label{ssec:bsq_kernel_choice}
As already noted, the posterior mean for the integral produced by the BSQ depends only on $\pi$ and the kernel controls the posterior variance of the integral.
The reasonableness of the BSQ output depends on the reasonableness of the assumption that $\nlFcnT*$ is ``well modelled'' by the GP specified by the kernel $k$.
Consequently, selection of the kernel is important in order to ensure that the integral variance is meaningful in modelling the integration error.
At the same time, the functional form of the kernel is constrained by the requirement in BSQ to analytically compute the integral of the kernel.
To facilitate analytic tractability of the Bayes--Sard moment transform, introduced next, we use the radial basis function (RBF) kernel
\begin{equation}\label{eq:kernel_rbf}
	\kerFcn(\inVar, \inVar') = \rbfScale^2 \prod_{\idInd=1}^{\inDim}\exp\pqty{ -\frac{  (\inVar*_\idInd - \inVar*_\idInd')^2 }{2\ell^2_\idInd} }
\end{equation}
throughout the remainder.
The parameters $\vb{\theta}$ of this kernel consist of the scale parameter $\alpha > 0$ and dimension-wise lengthscale parameters $\ell_1,\ldots,\ell_D > 0$.
A particular modelling assumption associated with this kernel is that the integrand is infinitely differentiable.
If this is not the case (i.e., there is model misspecification) the proposed method still works but the uncertainty quantification for the integral may be rendered less meaningful.
For certain classes of kernels it has been shown that convergence rates to the true integral as $N \to \infty$ are not much affected by model misspecification~\cite{Kanagawa2016,Kanagawa2017}.

\section{Bayes--Sard Moment Transform}\label{sec:Bayes--Sard_moment_transform}
The simplest way to design a moment transform is to use the BSQ directly for approximation of the moment integrals in \crefrange{eq:out_mean_approx}{eq:out_ccov_approx}.
However, this design does not reflect integral uncertainty, which is the key advantage of Bayesian quadrature, not to mention the fact that we would only obtain the classical rules as a result.
To resolve this issue, we employ the same general conceptual framework used in the design of the GPQ moment transform in \cite{Prueher2018}, which can account for the variance of the mean integral \labelcref{eq:out_mean_approx}.

\subsection{Incorporating Integration Error}\label{ssec:incorporating_integration_error}
First, it is important to realize that the output variable \( \outVar \) is now subject to an additional source of uncertainty in \( \nlFcn \) introduced by the model.
The key idea is to account for all sources of uncertainty in the computed moments, which can be achieved with the following
\begin{align}
	\outMean 	&= \E[\inVar]{\nlFcnT(\inVar)} 						& &\approx & &\outMeanApp 	= \E[\inVar,\,\nlFcn\mid\D]{\nlFcn(\inVar)} \\
	\outCov 	&= \Cov[\inVar]{\nlFcnT(\inVar), \nlFcnT(\inVar)} 	& &\approx & &\outCovApp 	= \Cov[\inVar,\,\nlFcn\mid\D]{\nlFcn(\inVar),\, \nlFcn(\inVar)} \\
	\inoutCov	&= \Cov[\inVar]{\inVar, \nlFcnT(\inVar)} 			& &\approx & &\inoutCovApp 	= \Cov[\inVar,\,\nlFcn\mid\D]{\inVar,\, \nlFcn(\inVar)}
\end{align}
Using the law of total expectation and covariance, the approximate moments of the output can be written as
\begin{align}
\outMeanApp 	&= \E[\nlFcn\mid\D]{\E[\inVar]{\nlFcn(\inVar)}} = \E[\inVar]{\E[\nlFcn\mid\D]{\nlFcn(\inVar)}}, \label{eq:out_mean_approx_decomp} \\
\outCovApp 		&= \Cov[\nlFcn\mid\D]{\E[\inVar]{\nlFcn(\inVar)}} + \E[\nlFcn\mid\D]{\Cov[\inVar]{\nlFcn(\inVar),\nlFcn(\inVar)}}, \label{eq:out_cov_approx_decomp_0} \\
				&= \Cov[\inVar]{\E[\nlFcn\mid\D]{\nlFcn(\inVar)}} + \E[\inVar]{\Cov[\nlFcn\mid\D]{\nlFcn(\inVar),\nlFcn(\inVar)}}, \label{eq:out_cov_approx_decomp_1} \\
\inoutCovApp	&= \E[\inVar]{\inVar\,\E[\nlFcn\mid\D]{\nlFcn(\inVar)}} - \E[\inVar]{\inVar}\E[\nlFcn\mid\D, \inVar]{\nlFcn(\inVar)}. \label{eq:out_ccov_approx_decomp}
\end{align}
The first equality exposes the fact that integral mean is obtained by integrating the mean function of the integrand model.
The way the integral uncertainty is incorporated into the output covariance is revealed by \cref{eq:out_cov_approx_decomp_0}.
Note that since the model of the integrand has conditionally independent outputs, the covariance of the integral, \( \Cov[\nlFcn\mid\D]{\E[\inVar]{\nlFcn(\inVar)}} \), and the model covariance, \( \Cov[\nlFcn\mid\D]{\nlFcn(\inVar),\nlFcn(\inVar)} \), are diagonal matrices.
When either of the covariances approaches zero, \crefrange{eq:out_mean_approx_decomp}{eq:out_ccov_approx_decomp} approach their true values.
From now on, we will work with the output covariance in the form \labelcref{eq:out_cov_approx_decomp_1} because it is easier to analyse and implement.

\subsection{Derivation of Transformed Moments}\label{ssec:bsqmt_derivation_transformed_moments}
In the following derivations, explicit conditioning on \( \D \) in the expectations is omitted to reduce notational clutter. We also assume that the stochastic decoupling substitution has taken place in the integrals, so that \( \nlFcnU(\inVarU) = \nlFcn(\inMean + \inCovFct\inVarU) \).


Taking the expression for the mean function of the model in \cref{eq:bsgp_multi-output_posterior_mean} and plugging it into \cref{eq:out_mean_approx_decomp}, the output mean of the Bayes--Sard quadrature moment transform (BSQMT) becomes
\begin{equation}
	\outMeanApp = \E[\inVarU]{\E[\nlFcn]{\nlFcnU(\inVarU)}} = \fevMat\T\vanMat\IT\E[\inVarU]{\vanVec[\inVarU]} = \fevMat\T\mtWm,
\end{equation}
where \( \mtWm = \vanMat\IT\E[\inVarU]{\vanVec[\inVarU]} \) are the mean weights.
The output covariance becomes
\begin{align}
	\outCovApp 
	&= \E[\inVarU]\big{\E[\nlFcn]{\nlFcnU(\inVarU)}\,\E[\nlFcn]{\nlFcnU(\inVarU)}\T} - \outMeanApp\outMeanApp\T + \E[\inVarU]{\Cov[\nlFcn]{\nlFcnU(\inVarU), \nlFcnU(\inVarU)}} \nonumber \\
	&= \fevMat\T\vanMat\IT\E[\inVarU]{\vanVec[\inVarU]\vanVec[\inVarU]\T}\vanMat\I\fevMat - \outMeanApp\outMeanApp\T + \gpExpVar\eye_\outDim \nonumber \\
	&= \fevMat\T\mtWc\fevMat - \outMeanApp\outMeanApp\T + \gpExpVar\eye_\outDim
\end{align}
where the expected model variance is
\begin{equation}\label{eq:expected_model_variance_gp}
	\gpExpVar = \E[\inVarU]{\kerFcn(\inVarU,\, \inVarU)} - \tr\bqty\big{\kervanCCov\T\vanMat\IT + \kervanCCov\vanMat\I - \mtWc\kerMat}.
\end{equation}
Here \( \kervanCCov = \E[\inVarU]{\kerVec[\inVarU]\vanVec[\inVarU]\T} \) and the covariance weights are \( \mtWc = \vanMat\IT\E[\inVarU]{\vanVec[\inVarU]\vanVec[\inVarU]\T}\vanMat\I \).
Finally, the covariance between the input and output becomes
\begin{align}
	\inoutCovApp 
	&= \E[\inVarU]{(\inMean + \inCovFct\inVarU)\, \E[\nlFcn]{\nlFcnU(\inVarU)}} - \E[\inVarU]{\inMean + \inCovFct\inVarU}\,\E[\nlFcn, \inVarU]{\nlFcnU(\inVarU)} \nonumber \\
	&= \inMean\outMeanApp + \inCovFct\E[\inVarU]{\inVarU\,\E[\nlFcn]{\nlFcnU(\inVarU)}} - \inMean\outMeanApp \nonumber \\
	&= \inCovFct\E[\inVarU]{\inVarU\,\vanVec[\inVarU]}\vanMat\I\fevMat = \inCovFct\mtWcc\fevMat
\end{align}
where the cross-covariance weights are \( \mtWcc = \E[\inVarU]{\inVarU\,\vanVec[\inVarU]}\vanMat\I \).

It has now become evident that the output moments depend on the expectations of the basis functions.
In \Cref{sec:Bayes--Sard_quadrature}, we have shown that the classical moment transforms can be recovered when the basis functions are multivariate polynomials.
When this basis and the RBF kernel \cref{eq:kernel_rbf} are used, the expectations above are available in closed form. 
Derivations are confined to \Crefrange{ssec:appendix_expectation_vv}{ssec:appendix_expectation_xv} so as not to disrupt the flow.
The complete algorithm of the Bayes--Sard moment transform is summarized in \Cref{alg:mt_bsq}.

\begin{algorithm}[]
	\DontPrintSemicolon
	\KwIn{The mean \( \inMean \) and the covariance \( \inCov \) of the input variable \( \inVar \), the integrand \( \nlFcn(\inVar) \), the matrix of unit sigma-points \( \spMatU \) and the kernel parameters \( \kerPar \).}
	\KwOut{Approximate mean \( \outMeanApp \) and covariance \( \outCovApp \) of the output variable \( \outVar = \nlFcn(\inVar) \), and approximate input-output covariance \( \inoutCovApp \).}
	\LinesNumbered
	\SetKwProg{Fn}{Function}{}{end}
	\SetKw{Return}{return}
	\SetKwFunction{BSQMT}{BSQMT}
	\SetKwFunction{chol}{MatrixFactor}
	
	\Fn{\BSQMT{\( \nlFcn(\inVar) \), \( \inMean \), \( \inCov \), \( \spMatU \), \( \kerPar \)}}{
		
		\BlankLine
		\tcp{form sigma-points}
		\( \inCovFct 	\leftarrow \) \chol{\( \inCov \)}\;
		\( \spMat 		\leftarrow \inMean + \inCovFct\spMatU \)\;
		
		\BlankLine
		\setstretch{1.1}
		\( \bar{k} 	\leftarrow	\E[\inVarU]{\kerFcn(\inVarU, \inVarU \,;\kerPar)} \)\; \label{ln:bsqmt_k_bar}
		\( \vanMean \leftarrow 	\E[\inVarU]{\vanVec[\inVarU]} \)\;
		\( \vanCov 	\leftarrow 	\E[\inVarU]{\vanVec[\inVarU]\vanVec[\inVarU]\T} \)\;
		\( \vanCCov \leftarrow 	\E[\inVarU]{\inVarU\vanVec[\inVarU]\T} \)\;
		\For{\( \spInd \leftarrow 1 \) \KwTo \( \spNum \)}{
			\( \bqty{\fevMat}_{\spInd *} 			\leftarrow \nlFcn(\inVar^{(\spInd)}) \)\;
			\( \bqty{\vanMat}_{\spInd *} 		\leftarrow \bmqty{\piFcn_1(\inVarU^{(\spInd)}) & \ldots & \piFcn_\spNum(\inVarU^{(\spInd)})} \)\;
			\( \bqty{\kervanCCov}_{\spInd *} 	\leftarrow \E[\inVarU]{\kerFcn(\inVarU, \inVarU^{(\spInd)}\,;\kerPar) \vanVec[\inVarU]\T} \)\; \label{ln:bsqmt_d}
			\For{\( m \leftarrow 1 \) \KwTo \( \spNum \)}{
				\( \bqty{\kerMat}_{\spInd m}	\leftarrow \kerFcn(\inVarU^{(\spInd)},\, \inVarU^{(m)};\, \kerPar) \)\;
			} \label{ln:bsqmt_end_for_k}
		}
		\BlankLine
		\tcp{compute BSQ weights}
		\( \mtWm	\leftarrow \vanMat\IT\vanMean \)\;
		\( \mtWc 	\leftarrow \vanMat\IT\vanCov\vanMat\I \)\;
		\( \mtWcc 	\leftarrow \vanCCov\vanMat\I \)\;
		
		\BlankLine
		\tcp{compute transformed moments}
		\( \gpExpVar 	\leftarrow \bar{k} - \tr\bqty\big{\kervanCCov\T\vanMat\IT + \kervanCCov\vanMat\I - \mtWc\kerMat} \)\;
		\( \outMeanApp 	\leftarrow \fevMat\T\mtWm \)\;
		\( \outCovApp 	\leftarrow \fevMat\T\mtWc\fevMat - \outMeanApp\outMeanApp\T + \gpExpVar\eye \)\;
		\( \inoutCovApp	\leftarrow \inCovFct\mtWcc\fevMat \)\;
		
		\BlankLine
		\Return \( \outMeanApp \), \( \outCovApp \), \( \inoutCovApp \)
	}
	\caption{
		Bayes-Sard quadrature moment transform 
	}
	\label{alg:mt_bsq}
\end{algorithm}

\begin{thm} 
	The BSQ output covariance \( \outCovApp \) is positive semi-definite.
\end{thm}
\begin{IEEEproof}
	Using the expression for the BSQ mean weights from \Cref{alg:mt_bsq}, we can write the output covariance as
	\( \outCovApp = \fevMat\T\vanMat\IT(\vanCov - \vanMean\vanMean\T)\vanMat\I\fevMat + \gpExpVar\eye_\outDim \).
	Define \( \mb{Z} = \vanMat\I\fevMat \) and \( \tilde{\vanCov} = \vanCov - \vanMean\vanMean\T \), then \( \outCov = \mb{Z}\T\tilde{\vanCov}\mb{Z} + \gpExpVar\eye_\outDim \).
	We recognize that \( \vphantom{\Big(} \tilde{\vanCov} = \V{\vanVec[\inVarU]} = \E{\vanVec[\inVarU]\vanVec[\inVarU]\T} - \E{\vanVec[\inVarU]}\E{\vanVec[\inVarU]}\T \succeq 0 \), which follows from the properties of covariance matrices.
	This implies that \( \mb{Z}\T\tilde{\vanCov}\mb{Z} \succeq 0 \) for any matrix \( \mb{Z} \).
	Because \( \gpExpVar \geq 0 \), we have that \( \outCovApp \succeq 0 \).
\end{IEEEproof}


\subsection{Relationship to the Gaussian Process Quadrature MT}\label{ssec:bsqmt_relationship_gpqmt}
The recently proposed Gaussian process quadrature moment transform (GPQMT) \cite{Prueher2018}, together with the BSQMT, are both instances of the general BQ framework.
The GPQMT uses a zero-mean GP prior model of the integrand as opposed to the more sophisticated hierarchical prior in \cref{eq:bsq_hierarchical_bsgp_prior_start,eq:bsq_hierarchical_bsgp_prior_end}.
As a result, the GPQMT weights are affected by the choice of kernel and its parameter values, which is not the case in the BSQMT, where the kernel only affects the last term of the transformed covariance and the weights depend only on the sigma-points and the choice of the function space \( \pi \).
Consequently, this makes BSQMT much less sensitive to misspecification of the kernel parameters, which is a notorious problem plaguing GPQMT.
\changed{Discussion of the choice of kernel parameters can be found in the original publication \cite{Prueher2018}.}

Compared to the zero-mean GP employed in GPQMT, the Bayes--Sard GP is a stronger prior, which means it can provide better fit to the integrand when conditioned on smaller datasets, such as the UT sigma-points, which are especially attractive in local filtering applications.

\subsection{BSQ Moment Transform in Sigma-Point Filtering}\label{ssec:bsmqt_filtering}
As outlined in \Cref{sec:sigma-point_filtering}, the local filtering algorithms use the moment transformations for computing the predictive moments of the system state and measurement.
\Cref{alg:bsq_kalman_filter} summarizes the Bayes--Sard quadrature Kalman filter (BSQKF), which employs the proposed BSQ moment transform for computing the predictive moments from \Cref{tab:moments}.
The BSQKF takes two different kernel parameter values, \( \kerPar_f \) and \( \kerPar_h \), because there are two different functions that need to be integrated (see \cref{eq:ssm_dynamics,eq:ssm_measurement_model}).

\begin{algorithm}[]
	\DontPrintSemicolon
	\KwIn{Sequence of measurements \( \Bqty\big{\msVar_{\tind}}_{\tind=1}^\Tind \), initial conditions \( \filtMean[0],\, \filtCov[0] \), kernel parameters \( \kerPar_{f} \) and \( \kerPar_{h} \), unit sigma-points \( \spMatU \)}
	\KwOut{Sequence of state estimates and covariances \( \Bqty\big{\filtMean,\, \filtCov}_{\tind=1}^\Tind \)}
	
	\BlankLine
	\SetKwFunction{BSQMT}{BSQMT}

	\For{\( k \leftarrow 1 \) \KwTo \( \Tind \)}{
		\setstretch{1.2}
		\tcp{predictive state moments} 
		\( \stMean_{\tind|\tind-1},\ \stCov_{\tind|\tind-1} \leftarrow \) \\ 
		\BSQMT{\( \stFcn(\inVar_{\tind-1}) \), \( \filtMean[\tind-1],\ \filtCov[\tind-1],\ \spMatU,\ \kerPar_{f} \)}\;
		\( \stCov_{\tind|\tind-1} \leftarrow \stCov_{\tind|\tind-1} + \stnCov \)\;
		
		\BlankLine
		\tcp{predictive measurement moments}
		\( \msMean_{\tind|\tind-1},\ \msCov_{\tind|\tind-1},\ \stMsCov_{\tind|\tind-1} \leftarrow \) \\ 
		\BSQMT{\( \msFcn(\inVar_{\tind}) \), \( \stMean_{\tind|\tind-1},\ \stCov_{\tind|\tind-1},\ \spMatU,\ \kerPar_{h} \)}\;
		\( \msCov_{\tind|\tind-1} \leftarrow \msCov_{\tind|\tind-1} + \msnCov \)\;
		
		\BlankLine
		\tcp{measurement update (filtering)}
		\( \kGain_\tind \leftarrow \stMsCov_{\tind|\tind-1}\pqty\big{\msCov_{\tind|\tind-1}}\I \)\;
		\( \filtMean \leftarrow \stMean_{\tind|\tind-1} + \kGain_\tind\pqty\big{\msVar_\tind - \msMean_\tind}\T \)\;
		\( \filtCov \leftarrow \stCov_{\tind|\tind-1} - \kGain_\tind\msCov_{\tind|\tind-1}\kGain_\tind\T \)\;
	}
	\caption{Bayes-Sard quadrature Kalman filter.}
	\label{alg:bsq_kalman_filter}
\end{algorithm}

\subsection{Kernel-Agnostic BSQ Moment Transform}\label{ssec:bsqmt_kernel-agnostic_bsqmt}
The BSQ moment transform stated in \Cref{alg:mt_bsq} depends on the choice of kernel and the values of its parameters, which in general GP regression are conventionally estimated from data by marginal likelihood maximization \cite{Rasmussen2006}.
Considering the fact that the classical sigma-point sets are inherently sparse relative to dimensionality of the input space, maximum likelihood estimation is not expected to work well in this context.
The last resort in practice is therefore manual parameter tuning.
When the moment transform is employed on its own, like in \Cref{ssec:mapping_polar_cartesian}, tuning can be informed by the analytic form of integrand with the goal of producing good model fit, which typically yields good results.
However, this tactic has not been observed to work well in filtering applications (see \Cref{ssec:univariate_growth_model,ssec:radar_tracking_reentry_vehicle}), because good filter performance may not be necessarily achieved by parameters that maximize model fit to the integrands \( \stFcn \) and \( \msFcn \) from \cref{eq:ssm_dynamics,eq:ssm_measurement_model}.

With all these considerations in mind, in filtering applications it is easier to re-parametrise the BSQMT in terms of the expected model variance (EMV) \( \gpExpVar \) directly rather than through the complicated dependence on kernel parameters as given by \cref{eq:expected_model_variance_gp}.
Note that this re-parametrisation renders the BSQMT independent of the choice of kernel and thus removes the need for computation of the EMV-related terms in \cref{ln:bsqmt_k_bar} and \crefrange{ln:bsqmt_d}{ln:bsqmt_end_for_k} in \Cref{alg:mt_bsq}. 

A more systematic approach than manual tuning would be to treat \( \gpExpVar \) as an unknown static parameter, which could be estimated using the energy function minimization \cite{Saerkkae2013}.
In case \( \gpExpVar \) is assumed to be time-variant, the state augmentation approach \cite{Haykin2001} might prove useful.
Exploration of the various parameter estimation techniques is beyond the scope of this article.

\section{Numerical Experiments}\label{sec:numerical_experiments}

This section contains a static example and two numerical filtering experiments that demonstrate performance of the Bayes--Sard quadrature moment transform and the Bayes--Sard quadrature Kalman filter.

\subsection{Mapping from Polar to Cartesian Coordinates}\label{ssec:mapping_polar_cartesian}
To gain insight into the performance of our proposed method, we first consider the moment transform in isolation, outside of the filtering context.
As an example of ubiquitous nonlinearity, we use the conversion from polar to Cartesian coordinates, which appears in radars and laser range finders and is given by
\begin{equation}\label{eq:polar2cartesian}
	\bmqty{x \\ y} = \bmqty{r\cos(\theta) \\ r\sin(\theta)},
\end{equation}
where \( r \) is radius and \( \theta \) is azimuth.
When setting the RBF kernel parameters, we exploited the conditional linearity of the mapping, which is reflected in the setting $\ell_1 = 60$ and $\ell_2 = 6$, where the range coordinate \( r \) is given relatively large value.
The kernel scaling was set to \( \alpha=1 \).

The BSQ and GPQ transforms with UT points (\( \kappa = 2 \)) were compared with the classical unscented transform. 
Since the input space is given in polar coordinates, the input means were placed on a spiral and several covariances were considered for each input mean.
More specifically, 10 different positions on a spiral were chosen as input means \( \inMean_i = \bmqty{r_i & \theta_i} \) and for each \( \inMean_i \) we considered 10 different input covariance matrices \( \inCov_j = \diag{\bmqty{\sigma^2_r & \sigma^2_{\theta,j}}} \), where \( \sigma_r = \SI{0.5}{m} \) and \( \sigma_{\theta,j} \in \bqty{\SI{6}{\degree},\, \SI{36}{\degree}} \) for \( j = 1, \ldots, 10 \).
Agreement between the approximate moments \( (\outMeanApp,\ \outCovApp) \) and the ground-truth moments \( (\outMean,\ \outCov) \), computed by Monte Carlo method with \num{10000} samples, was measured by the symmetrized KL-divergence\footnote{The advantage of SKL over KL-divergence is that it is symmetric; that is, it holds \( \SKL{\outMean, \outCov}{\outMeanApp, \outCovApp} = \SKL{\outMeanApp, \outCovApp}{\outMean, \outCov} \).} of two Gaussian densities given by 
\begin{align}\label{eq:skl}
\SKL{\outMeanApp, \outCovApp}{\outMean, \outCov}
&= \frac{1}{4}\left\{(\outMean - \outMeanApp)\T\outCov\I(\outMean-\outMeanApp) + \right. \nonumber \\
&\mathrel{\phantom{=}} \phantom{\frac{}{4}\left\{\right.} (\outMeanApp - \outMean)\T\outCovApp\I(\outMeanApp - \outMean) + \nonumber \\
&\mathrel{\phantom{=}} \phantom{\frac{}{4}\left\{\right.} \left. \trace(\outCov\I\outCovApp) + \trace(\outCovApp\I\outCov) - 2\outDim\right\},
\end{align}
where \( E = \dim(\outMean) \).

\begin{figure}[!h]
	\centering
	\input{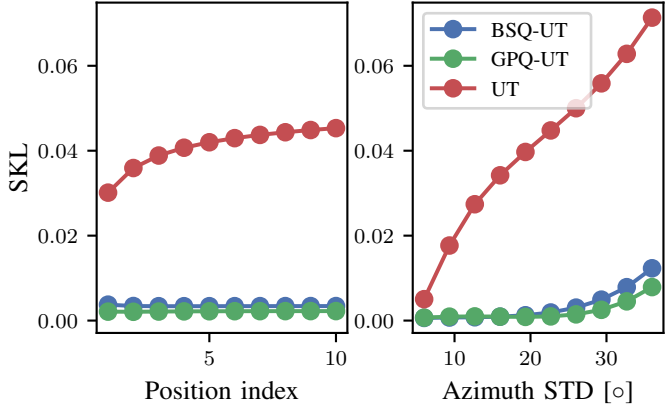}
	\caption{Symmetrised KL divergence between the MC baseline Gaussian and the approximate Gaussian computed by the BSQ, GP and UT moment transforms.}
	\label{fig:polar2cartesian_skl}
\end{figure}

The plots in the left pane of \Cref{fig:polar2cartesian_skl} show the SKL score averaged over the input covariances \( \inCov_j \), while the right pane shows SKL averaged over the input means \( \inMean_i \).
The proposed Bayes--Sard quadrature transform with UT points (BSQ-UT) outperforms the classical UT, while achieving almost the same performance as the GPQ-UT.
\changed{What is especially noteworthy here, is the fact that BSQ-UT computes the transformed mean $ \outMeanApp $ identically to the UT and the only difference is in the manner in which the transformed covariance $ \outCovApp $ is computed.
The \Cref{fig:polar2cartesian_skl} shows that the BSQ-UT, which can be viewed as a probabilistic equivalent of the UT, offers improved performance, because it accounts for the numerical integration error of the UT.
For completeness we also included the GPQ-UT transform, which dominates this example, but does not have such a clear theoretical connection to the UT.}

\subsection{Univariate Non-stationary Growth Model}\label{ssec:univariate_growth_model}
As the first experiment with sigma-point filters based on the BSQ moment transform, we consider the univariate non-stationary growth model (UNGM), which is often used to benchmark particle filters \cite{Gordon1993,Kitagawa1996}.
The system dynamics and the observation model are given by
\begin{align}
	\stVar*_\tind	&= \frac{1}{2}\stVar*_{\tind-1} + \frac{25\stVar*_{\tind-1}}{1+{\stVar*}^2_{\tind-1}} + 8\cos(1.2\,\tind) + \stnVar*_{\tind-1}, \\
	\msVar*_\tind 	&= \frac{1}{20}{\stVar*}^2_{\tind-1} + \msnVar*_\tind,
\end{align}
with the state noise \( \N*[\stnVar*_{\tind-1}][0][10] \), measurement noise \( \N*[\msnVar*_\tind][0][1] \) and initial conditions \( \stVar*_0 = \N*[\stVar*_{0|0}][0][5] \).
Kernel scaling used in the BSQ with for the UT (\( \kappa = 2 \)) and the 7-th order GH points was set to \( \alpha = 3 \) and the lengthscales to \( \ell = 0.3 \) and \(\ell = 0.4 \), respectively.
For the 5-th order GH points the kernel parameters were set to \( \alpha=5 \) and \( \ell = 0.6 \).

The RMSE was used to measure the tracking performance.
The inclination indication (INC) \cite{Li2006}, given by
\begin{equation}\label{eq:inclination_indicator}
\mathrm{INC} = \frac{10}{\Tind}\sum_{\tind=1}^{\Tind}\log_{10} \frac{(\stVar_\tind - \stMean_{\tind|\tind})\T(\stCov_{\tind|\tind})\I(\stVar_\tind - \stMean_{\tind|\tind})}{(\stVar_\tind - \stMean_{\tind|\tind})\T\mb{\Sigma}_\tind\I(\stVar_\tind - \stMean_{\tind|\tind})},
\end{equation}
where \( \mb{\Sigma}_\tind \) is the mean-squared error matrix of the state, was used to measure the credibility of the estimates. 
A perfectly balanced estimate has \( \mathrm{INC} = 0 \). 
For \( \mathrm{INC} > 0 \), the estimate is said to be optimistic, which is to say the covariance is smaller than it should be, while negative values indicate pessimism.
\changed{Refer to \cite{Li2006} for other related credibility measures.}
We simulated the model for \( \Tind = 500 \) time steps and averaged the performance scores over \( 100 \) simulations.
The variance of the average scores was estimated by bootstrapping. 
The parentheses in \Cref{tab:ungm_rmse,tab:ungm_inc} contain the uncertainty as the least significant digits of 2 standard deviations.

The BSQ filters with classical points were tested against the well-known sigma-point filters as well as the GPQ filters from \cite{Prueher2018}.
As seen in \Cref{tab:ungm_rmse}, the filters based on the BSQ outperform the classical sigma-point filters in terms of RMSE. 
Assuming the GH points are used, BSQKFs can outperform the GPQ filters as well.
In comparison with the classical filters, the proposed BSQ filters also provide much more balanced estimates as evidenced by the values of the INC in \Cref{tab:ungm_inc}.

\begin{table}[h!]
\centering
\begin{tabular}{@{} l l l l @{}}
	\toprule
	{} 	 & Classical 			  & GPQ 	   	   		   & BSQ         \\
	\midrule
	UT   & 			 10.81 (0.14) & 		  10.37 (0.08) & 9.70 (0.12) \\
	GH-5 & 			 10.03 (0.14) & \phantom{0}9.01 (0.11) & 8.82 (0.08) \\
	GH-7 & \phantom{0}9.74 (0.13) & \phantom{0}8.80 (0.10) & 8.61 (0.09) \\
	\bottomrule
\end{tabular}
\caption{Filter RMSE for the UNGM example.}
\label{tab:ungm_rmse}
\end{table}

\begin{table}[h!]
\centering
\begin{tabular}{@{} l l l l @{}}
	\toprule
	{} 	 & Classical    		  & GPQ 	    & BSQ 	  	 \\
	\midrule
	UT   & 			 12.17 (0.06) & 4.87 (0.01) & 4.57 (0.03)  \\
	GH-5 & 			 10.33 (0.07) & 5.26 (0.03) & 1.85 (0.02)  \\
	GH-7 & \phantom{0}9.27 (0.07) & 4.95 (0.03) & 2.52 (0.03)  \\
	\bottomrule
\end{tabular}
\caption{Filter INC for the UNGM example.}
\label{tab:ungm_inc}
\end{table}

\subsection{Radar Tracking of Reentry Vehicle}\label{ssec:radar_tracking_reentry_vehicle}
Radar tracking of an object entering Earth's atmosphere is a very challenging tracking scenario for nonlinear filters.
The following model has been used to demonstrate superiority of the UKF over the EKF in \cite{Julier2004}.

The dataset of the ground-truth system state trajectories comprised of \num{100} MC simulations of the SDE
\begin{equation}\label{eq:reentry_vehicle_system}
	\dot{\stVar}(t) = 
	\bmqty{
		v^x(t) \\ 
		v^y(t) \\ 
		D(t) v^x(t) + G(t) p^x(t) + \stnVar*_{1}(t) \\ 
		D(t) v^y(t) + G(t) p^y(t) + \stnVar*_{2}(t) \\
		\stnVar*_{3}(t)
	}
\end{equation}
using the Euler-Maruyama method for the duration of \SI{200}{s} with time step \( \Delta t = \SI{0.05}{s} \).
The state \( \stVar(t) = \bmqty{p^x(t) & p^y(t) & v^x(t) & v^y(t) & \theta(t)} \) consists of Cartesian position coordinates \( (p^x, p^y) \) in \si{km}, velocity \( (v^x, v^y) \) in \si{km/s} and an aerodynamic parameter \( \theta \).
The functions \( D(t) \) and \( G(t) \) are the drag-related and the gravity-related force terms, given by
\begin{align}
	D(t) 	&= \beta(t) \exp\pqty{\frac{R_0 - R(t)}{H_0}}V(t), \\
	G(t) 	&= \frac{G m_0}{r^3(t)}, \\
	\beta(t) &= \beta_0 \exp(\theta(t)),
\end{align}
with \( R(t) = \sqrt{(p^x(t))^2 + (p^y(t))^2} \) being the distance from the centre of the Earth and \( V(t) = \sqrt{(v^x(t))^2 + (v^y(t))^2} \) the speed.
The parameter values were set identically to \cite{Julier2004}; thus \( R_0 = 6374 \), \( H_0 = 13.406 \), \( \beta_0 = \num{-0.59783} \) and \( Gm_0 = \num{3.9860e5} \).
The system initial condition \( \N*[\stVar(0)][\stMean_0][\stCov_0] \) was set to
\begin{align}
	\stMean_0 &= \bmqty{6500 & 350 & -1.8 & -6.8 & 0.7}, \label{eq:reentry_vehicle_system_init_mean} \\
	\stCov_0  &= \diag{\bmqty{\sigma^2_p & \sigma^2_p & \sigma^2_v & \sigma^2_v & 0}}, \label{eq:reentry_vehicle_system_init_covariance}
\end{align}
where \( \sigma^2_p = \SI{1e-6}{km^{2}} \) and \( \sigma^2_v = \SI{1e-6}{km^2.s^{-2}} \).
The vector \( \stnVar(t) = \bmqty{q_{1}(t) & q_{2}(t) & q_{3}(t)} \) is the white noise process with covariance function 
\begin{equation}\label{}
	\Cov{\stnVar(t), \stnVar(s)} = \delta(t-s)\diag{\bmqty{\sigma^2_v & \sigma^2_v & \sigma^2_\theta}}
\end{equation}
where \( \delta(\cdot) \) is the Dirac delta distribution and \( \sigma^2_v = \SI{2.4e-5}{km^2.s^{-2}} \) and \( \sigma^2_\theta = \num{0} \).
For every generated trajectory we simulated the range and bearing measurements of the radar using the measurement model
\begin{equation}\label{eq:radar_measurement_model}
	\msVar_\tind = \bmqty{\sqrt{\vphantom{\big(} (p^x_\tind)^2 + (p^y_\tind)^2} \\ \atantwo(p^y_\tind,\, p^x_\tind)} + \msnVar_\tind
\end{equation}
with the measurement noise covariance 
\begin{equation}
\msnCov = \diag{\bqty{\mqty{\SI{1e-6}{km^2} & \SI{0.17e-6}{rad^2} }}}.
\end{equation}

The filters assumed the discrete-time state-space model
\begin{equation}\label{eq:reentry_vehicle_model}
	\stVar_\tind = \stVar_{\tind-1} + 
	\bmqty{
		v^x_\tind\, \Delta t \\ 
		v^y_\tind\, \Delta t \\ 
		\vphantom{\big(}(D_\tind v^x_\tind + G_\tind p^x_\tind)\, \Delta t + \stnVar*_{1,\tind} \\ 
		\vphantom{\big(}(D_\tind v^y_\tind + G_\tind p^y_\tind)\, \Delta t + \stnVar*_{2,\tind} \\
		\stnVar*_{3,\tind}
	}
\end{equation}
obtained by Euler-Maruyama approximation of \cref{eq:reentry_vehicle_system} with step size \( \Delta t = \SI{0.1}{s} \).
The resulting process covariance was \( \stnCov = \Delta t\cdot \diag{\bmqty{\sigma^2_v & \sigma^2_v & \sigma^2_\theta}} \) where we set \( \sigma^2_\theta = \num{1e-6} \).
We considered a scenario with misspecified initial conditions, where the filter assumed that the initial mean and covariance of the system are
\begin{align}\label{eq:radar_tracking_model_init}
	\filtMean[0] 	&= \bmqty{6500 & 350 & -1.1 & -6.1 & 0.7}, \\
	\filtCov[0]		&= \diag{\bmqty{\sigma^2_p & \sigma^2_p & \sigma^2_v & \sigma^2_v & 1}},
\end{align}
which, compared to the true conditions in \cref{eq:reentry_vehicle_system_init_mean,eq:reentry_vehicle_system_init_covariance}, indicates that the initial filtered covariance is not reflective of our ignorance about the true initial velocity.

In the experiments, we compared the BSQKF with UT sigma-points with the standard UKF. 
The previously proposed GPQKF is not included in the results, because we were unable to find kernel parameter values yielding numerically stable estimates.\footnote{\changed{Experimental evidence strongly suggests that the GPQKF is generally less suitable for problem where dimension of the state vector is > 3.}}
This is were the advantages of the BSQKF, described in \Cref{ssec:bsqmt_relationship_gpqmt}, become practically evident.
For this experiment, the BSQKF employed the kernel-agnostic BSQMT, described in \Cref{ssec:bsqmt_kernel-agnostic_bsqmt}, with \( \gpExpVar_{f} = \num{2e-4}\eye \) for the dynamics and to  \( \gpExpVar_{h} = \diag{\bmqty{0 & 0}} \) for the measurement model.
Both filters used identical UT sigma-points with scaling parameter \( \kappa = 0 \).

\begin{figure}[h]
	\centering
	\input{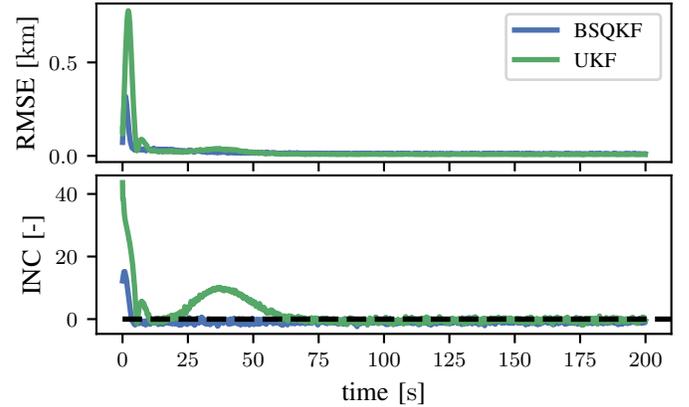}
	\caption{Position RMSE and INC.}
	\label{fig:radar_tracking_position}
\end{figure}

\begin{figure}[h]
	\centering
	\input{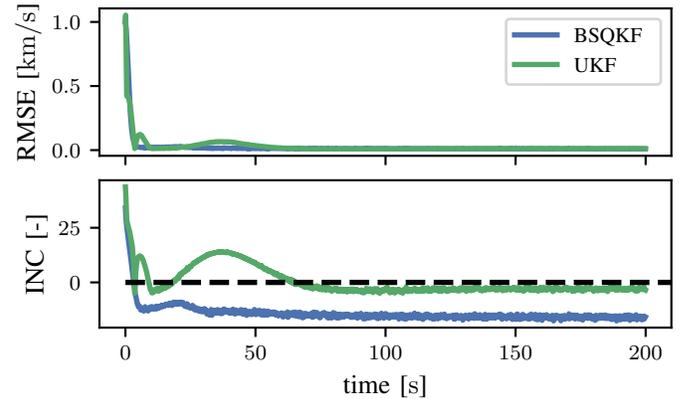}
	\caption{Velocity RMSE and INC.}
	\label{fig:radar_tracking_velocity}
\end{figure}

\begin{figure}[!h]
	\centering
	\input{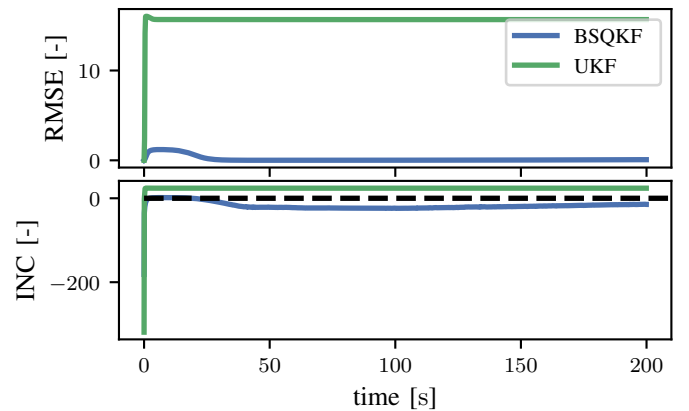}
	\caption{Aerodynamic parameter RMSE and INC.}
	\label{fig:radar_tracking_parameter}
\end{figure}

\Crefrange{fig:radar_tracking_position}{fig:radar_tracking_parameter} compare the development of the RMSE and INC in time for the position, the velocity and the aerodynamic parameter.
Overall, the proposed BSQKF clearly provides more credible estimates for the position and the aerodynamic parameter, especially during the initial stages, while the velocity estimates are more pessimistic, which is generally preferable to optimism in safety-critical applications.
In terms of RMSE, the BSQKF gives comparable estimate of the velocity and superior estimates of position and parameter.
Note, in \Cref{fig:radar_tracking_parameter}, how the parameter RMSE for the UKF stays constant, while for the BSQKF it is much lower.
\changed{\Cref{tab:reentry_summary} summarizes the results.}


\begin{table}[h!]
\centering
\begin{tabular}{@{} lrrrrr @{}}
	\toprule
			  					& \multicolumn{2}{c}{RMSE} 	  	  & &  \multicolumn{2}{c}{INC}   	  \\
	\cmidrule{2-3}\cmidrule{5-6}
	{} 		  					&  BSQKF 		  &  UKF 	      & &   BSQKF 	   	   &  UKF 	  	  \\
	\midrule
	Position [\si{km}]  		&          0.018 &        0.025 & &         -0.967 &       1.204 \\
	Velocity [\si{km/s}] 		&          0.020 &        0.026 & &        -14.464 &       0.045 \\
	Parameter [-] 				&          0.137 &       15.580 & &        -17.213 &      23.792 \\
	\bottomrule
\end{tabular}
\caption{RMSE and INC averaged over time and simulations for position, velocity and the aerodynamic parameter.}
\label{tab:reentry_summary}
\end{table}

\section{Conclusions and Discussion}\label{sec:conclusion}
In this article, we designed a general-purpose moment transformation based on Bayes--Sard quadrature, which allows for explicit modelling of numerical integration error through the use of a stochastic process model.
The hierarchical GP prior was shown to be key in developing probabilistic models whose means correspond to the classical quadrature rules used in the sigma-point filters and whose variance is statistically meaningful.
We designed the BSQ Kalman filter by leveraging the proposed BSQ moment transform for computation of the predictive moments.

The BSQMT could easily be applied in nonlinear sigma-point smoothing as well.
Further improvement in performance could be achieved by the use of more expressive multi-output GP models.

\appendix[Computation of Certain Expectations]
This appendix derives analytical expressions for a number of expectations that are needed for implementation of the Bayes--Sard moment transform.
In the following, let \( \N* \) and let \( \vanVec : \R^\inDim\to\R^\piDim \) consist of monomials defined by the multi-indices \( \mulInd_1,\, \ldots,\, \mulInd_\piDim \in \Nat_0^\inDim \), such that \( \vanVec = \bmqty{\inVar^{\mulInd_1} & \ldots & \inVar^{\mulInd_\piDim}}\T \).
Given a sigma-point set \( \spSet = \Bqty{\inVar_1,\, \ldots,\, \inVar_\spNum} \subset \R^\inDim \), let \( \kerVec \) denote a vector of kernel evaluations, such that \( \bqty{\kerVec}_\spInd = \kerFcn(\inVar, \inVar^{(\spInd)}) \), where the kernel is the RBF kernel in \cref{eq:kernel_rbf}.

\subsection{Computing \( \E{\vanVec\vanVec\T} \)}\label{ssec:appendix_expectation_vv}
Elements of the $\piDim \times \piDim$ matrix $\vanCov \triangleq \E{\vanVec\vanVec\T}$ are
\begin{equation*}
	[\vanCov]_{qr} = \prod_{\idInd=1}^\inDim \frac{1}{\sqrt{2\pi}} \int_\R\! \inVar*_\idInd^{\alpha_{q,\idInd}+\alpha_{r,\idInd}} \mathrm{e}^{-\inVar*_\idInd^2/2} \dx[\inVar*_\idInd].
\end{equation*}
From the standard Gaussian moment formula
\begin{equation*}
	\frac{1}{\sqrt{2\pi}} \int_{\R}\! \inVar*^n \mathrm{e}^{-\inVar*^2/2} \dx = \begin{cases} (n-1)!! & \text{if $n$ is even,} \\ 0 & \text{if $n$ is odd} \end{cases}
\end{equation*}
we obtain
\begin{equation*}
	[\vanCov]_{qr} = \prod_{\idInd=1}^\inDim (\alpha_{q,\idInd} + \alpha_{r,\idInd} - 1)!!
\end{equation*}
if none of $\alpha_{q,\idInd} + \alpha_{r,\idInd}$ for $\idInd = 1,\, \ldots,\, \inDim$ is odd and $[\vanCov]_{qr} = 0$ otherwise.

\subsection{Computing \( \E{\kerVec\vanVec\T} \)}\label{ssec:appendix_expectation_kv}
Elements of the $\spNum \times \piDim$ matrix $\kervanCCov \triangleq \E{\kerVec\vanVec\T}$ are
\begin{equation*}
	[\kervanCCov]_{\spInd\piInd} = \prod_{\idInd=1}^\inDim \frac{1}{\sqrt{2\pi}} \int_\R 
	\inVar*_\idInd^{\alpha_{\piInd,\idInd}} \exp\pqty{ -\frac{(\inVar*_\idInd - \inVar*_{\spInd,\idInd})^2}{2\ell_\idInd^2} } \mathrm{e}^{-\inVar*_\idInd^2/2} \dx[\inVar*_\idInd].
\end{equation*}
Completion of squares and a change of variables yield
\begin{equation*}
\begin{split}
\int_\R 
&x_\idInd^{\alpha_{\piInd,\idInd}} \exp\pqty{ \! -\frac{(x_\idInd - x_{\spInd,\idInd})^2}{2\ell_\idInd^2} } \mathrm{e}^{-x_\idInd^2/2} \dx[x_\idInd] \\
&= \int_\R x_\idInd^{\alpha_{\piInd,\idInd}} \exp\pqty{ \! -\frac{\pqty{x_\idInd - \frac{1}{1+\ell_\idInd^2} x_{\spInd,\idInd} }^2}{2\frac{\ell_\idInd^2}{1+\ell_\idInd^2}} } \\
& \hspace{2cm} \times \exp\pqty{ \! -\frac{x_{\spInd,\idInd}^2}{2(1+\ell_\idInd^2)} } \dx[x_\idInd] \\
&= \pqty{ \frac{\ell_\idInd^2}{1+\ell_\idInd^2} }^{1/2} \exp\pqty{ \! -\frac{x_{\spInd,\idInd}^2}{2(1+\ell_\idInd^2)} } \\
& \hspace{0.5cm} \times \int_\mathbb{R} \bqty{ \pqty{ \frac{\ell_\idInd^2}{1+\ell_\idInd^2} }^{1/2} x_\idInd + \frac{x_{\spInd,\idInd}}{1+\ell_\idInd^2}}^{\alpha_{\piInd,\idInd}} \mathrm{e}^{-x_\idInd^2/2} \dx[x_\idInd] \\
&= \ell_\idInd \pqty{ \frac{\ell_\idInd^2}{1+\ell_\idInd^2} }^{(1+\alpha_{\piInd,\idInd})/2} \exp\pqty{ \! -\frac{x_{\spInd,\idInd}^2}{2(1+\ell_\idInd^2)} } \\
& \hspace{0.5cm} \times \int_\R \pqty{ \ell_\idInd x_\idInd + \frac{x_{\spInd,\idInd}}{\sqrt{1+\ell_\idInd^2}} }^{\alpha_{\piInd,\idInd}} \mathrm{e}^{-x_\idInd^2/2} \dx[x_\idInd].
\end{split}
\end{equation*}
The binomial theorem yields
\begin{equation*}
\begin{split}
&\pqty{ \ell_\idInd x_\idInd + \frac{x_{\spInd,\idInd}}{\sqrt{1+\ell_\idInd^2}} }^{\alpha_{\piInd,\idInd}} \\
&\phantom{\bigg(\ell_\idInd x_\idInd}= \sum_{s=0}^{\alpha_{\piInd,\idInd}} \frac{\alpha_{\piInd,\idInd}!}{s!(\alpha_{\piInd,\idInd}-s)!} \ell_\idInd^s x_\idInd^s \pqty{ \frac{x_{\spInd,\idInd}}{\sqrt{1+\ell_\idInd^2}} }^{\alpha_{\piInd,\idInd}-s}.
\end{split}
\end{equation*}
Therefore
\begin{equation*}
\begin{split}
&\hspace{-0.2cm}\frac{1}{\sqrt{2\pi}} \int_\R \pqty{ \ell_\idInd x_\idInd + \frac{x_{\spInd,\idInd}}{\sqrt{1+\ell_\idInd^2)}} }^{\alpha_{\piInd,\idInd}} \mathrm{e}^{-x_\idInd^2/2} \dx[x_\idInd] \\
&= \sum_{s=0}^{\alpha_{\piInd,\idInd}} \frac{\alpha_{\piInd,\idInd}!}{s!(\alpha_{\piInd,\idInd}-s)!} \ell_\idInd^s \pqty{ \frac{x_{\spInd,\idInd}}{\sqrt{1+\ell_\idInd^2}} }^{\alpha_{\piInd,\idInd}-s} \\
& \hspace{1cm} \times \frac{1}{\sqrt{2\pi}} \int_\R x_\idInd^s \mathrm{e}^{-x_\idInd^2/2} \dx[x_\idInd] \\
&= \sum_{s=0}^{\lfloor \alpha_{\piInd,\idInd}/2 \rfloor } \frac{\alpha_{\piInd,\idInd}!}{(2s)!(\alpha_{\piInd,\idInd}-2s)!} \ell_\idInd^{2s} \pqty{ \frac{x_{\spInd,\idInd}}{\sqrt{1+\ell_\idInd^2}} }^{\alpha_{\piInd,\idInd}-2s} \frac{(2s)!}{2^s s!} \\
&= \sum_{s=0}^{\lfloor \alpha_{\piInd,\idInd}/2 \rfloor} \frac{\alpha_{\piInd,\idInd}!}{2^s s! (\alpha_{\piInd,\idInd}-2s)!} \ell_\idInd^{2s} \pqty{ \frac{x_{\spInd,\idInd}}{\sqrt{1+\ell_\idInd^2} }}^{\alpha_{\piInd,\idInd}-2s}.
\end{split}
\end{equation*}
By combining all the equations above, we conclude that
\begin{equation*}
\begin{split}
[\kervanCCov]_{\spInd\piInd} ={}& \prod_{\idInd=1}^\inDim \Bigg[ \ell_\idInd \bigg( \frac{\ell_\idInd^2}{1+\ell_\idInd^2} \bigg)^{(1+\alpha_{\piInd,\idInd})/2} \exp\bigg( \! -\frac{x_{\spInd,\idInd}^2}{2(1+\ell_\idInd^2)} \bigg) \Bigg. \\
& \times \Bigg. \sum_{s=0}^{\lfloor \alpha_{\piInd,\idInd}/2 \rfloor } \frac{\alpha_{\piInd,\idInd}!}{2^s s! (\alpha_{\piInd,\idInd}-2s)!} \ell_\idInd^{2s} \bigg( \frac{x_{\spInd,\idInd}}{\sqrt{1+\ell_\idInd^2}} \bigg)^{\alpha_{\piInd,\idInd}-2s} \Bigg].
\end{split}
\end{equation*}

\subsection{Computing \( \E{\inVar\vanVec} \)}\label{ssec:appendix_expectation_xv}
This reduces to the computation in \Cref{ssec:appendix_expectation_vv}. The $\inDim \times \piDim$ matrix $\vanCCov \triangleq \E{\inVar\vanVec\T}$ has the elements
\begin{equation*}
	\bqty{\vanCCov}_{\idInd\piInd} = \alpha_{\piInd,\idInd} \prod_{\idInd \neq \piInd} (\alpha_{\piInd,\idInd}-1)!!
\end{equation*}
if $\alpha_{\piInd,\idInd}$ is odd and none of $\alpha_{\piInd,\idInd}$ are and $\bqty{\vanCCov}_{\idInd\piInd} = 0$ otherwise.

\section*{Acknowledgment}
JP and OS were supported by the Czech Ministry of Education, Youth and Sports, project LO1506.
TK was supported by the Aalto ELEC Doctoral School. 
CJO was supported by the Lloyd's Register Foundation programme on Data-Centric Engineering at the Alan Turing Institute, UK.
SS was supported by the Academy of Finland project 313708.

\bibliographystyle{IEEEtran}
\bibliography{./refdb}

\begin{IEEEbiography}[{\includegraphics[width=1in,height=1.25in,clip,keepaspectratio]{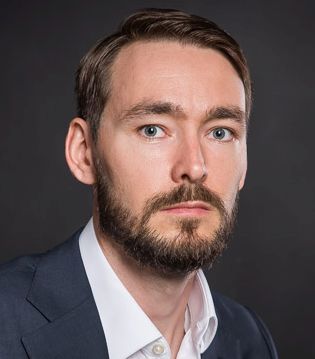}}]{Jakub Pr\"{u}her}
	is a Ph.D. student in the Identification and Decision Making research group at NTIS - New Technologies for the Information Society.
	He received his bachelor's degree in computer science in 2011 and his master's degree in cybernetics in 2013, both from the University of West Bohemia.
	His research interests include nonlinear state estimation methods, use of Bayesian reasoning in numerical algorithms and machine learning in general.
\end{IEEEbiography}
\begin{IEEEbiography}[{\includegraphics[width=1in,height=1.25in,clip,keepaspectratio]{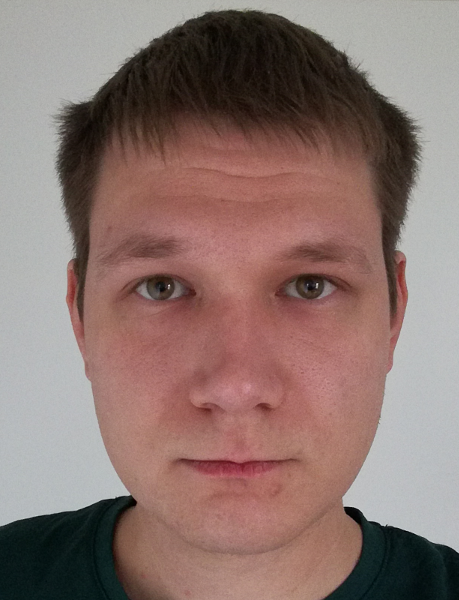}}]{Toni Karvonen}
received his Master of Science degree in applied mathematics from the University of Helsinki in 2015. Since 2016, he has been a doctoral student at the Department of Electrical Engineering and Automation, Aalto University, Finland. 
His research interests are in stochastic state estimation and numerical analysis, in particular numerical integration and probabilistic numerics.
\end{IEEEbiography}
\begin{IEEEbiography}[{\includegraphics[width=1in,height=1.25in,clip,keepaspectratio]{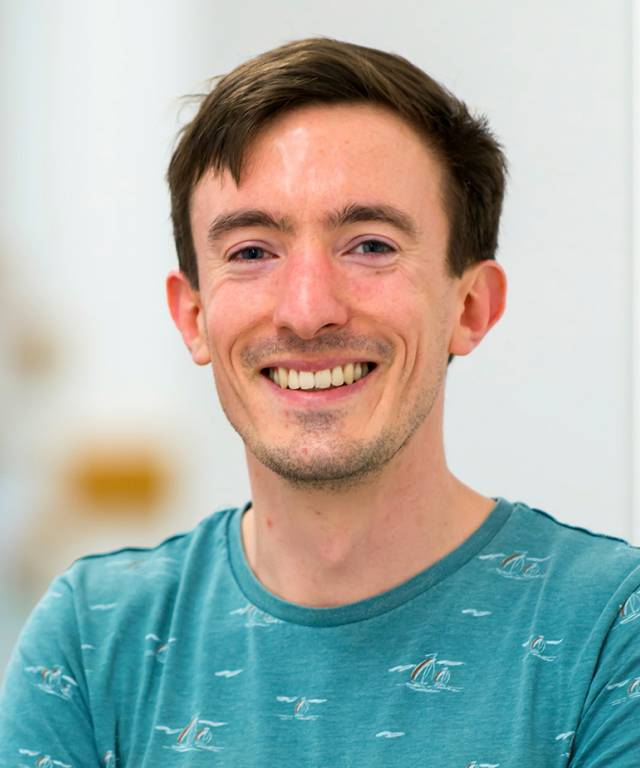}}]{Chris J. Oates}
 received his Ph.D. degree in statistics and complexity science from the University of Warwick, Coventry, UK, in 2013. 
 Since 2018, he has been a Professor in the School of Mathematics, Statistics and Physics, Newcastle University, Newcastle-upon-Tyne, UK. 
 He is also a Group Leader for the Lloyd's Register Foundation programme on Data-Centric Engineering at the Alan Turing Institute, London, UK, where he focuses on computational statistical techniques and the development of probabilistic numerical methods. 
 Prof. Oates was a recipient of the biennial Research Prize of the Royal Statistical Society in 2017.
\end{IEEEbiography}

\begin{IEEEbiography}[{\includegraphics[width=1in,height=1.25in,clip,keepaspectratio]{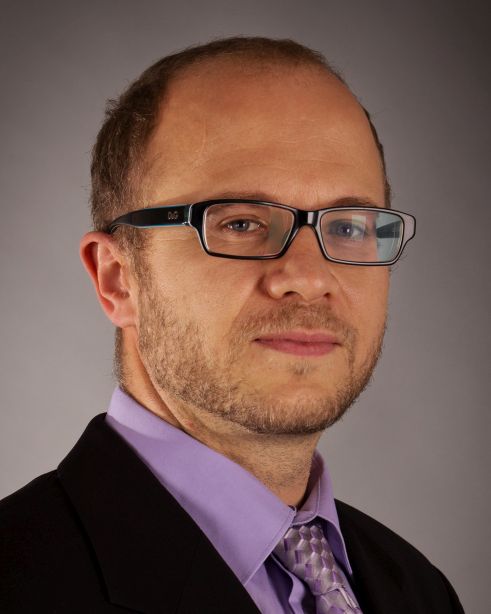}}]{Ond\v{r}ej Straka}
	(M’16) received the master’s degree in cybernetics and control engineering and the Ph.D. degree in cybernetics from the University of West 	Bohemia, Pilsen, Czech Republic, in 1998 and 2004, respectively. 
	Since 2015, he has been an Associate Professor with the Department of Cybernetics, University of West Bohemia. He is the Head of the Identification and Decision Making Research Group (IDM), NTIS—New Technologies for the Information Society, where he focuses on nonlinear state estimation, stochastic systems, and system identification. 
	He has participated in a number of projects of fundamental research and in several project of applied research (e.g., GNSS-based safe train localization and attitude and heading reference system). 
	He was involved in development of several software frameworks for nonlinear state estimation and system identification. 
	He has published over 70 journal and conference papers in journals, such as Automatica, the IEEE Transactions on Automatic Control, the IEEE Transactions on Aerospace and Electronic Systems, the IEEE Transactions on Cybernetics, and Signal Processing and at international conferences such as American Control Conference, World Congresses and Symposia of the IFAC, and FUSION Conferences. 
	His current research interests include local and global nonlinear state estimation methods, noise CM estimation in state-space models, performance evaluation, and fault detection in navigation systems. 
	Dr. Straka was a recipient of Werner von Siemens Excellence Award in 2014 for most important result in the basic research.
\end{IEEEbiography}

\begin{IEEEbiography}[{\includegraphics[width=1in,height=1.25in,clip,keepaspectratio]{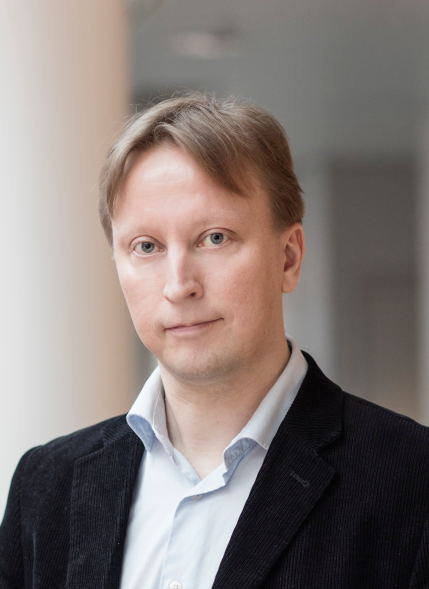}}]{Simo S\"{a}rkk\"{a}}
	received his Doctor of Science (Tech.) degree in electrical and communications engineering from Helsinki University of Technology, Espoo, Finland, in 2006. 
	From 2000 to 2010 he worked with Nokia Ltd., Indagon Ltd., and Nalco Company in various industrial research projects related to telecommunications, positioning systems, and industrial process control. 
	From 2010 to 2015 he worked as a Senior Researcher with the Department of Biomedical Engineering and Computational Science (BECS) of Aalto University, Finland. 
	Currently, he is an Associate Professor with the Department of Electrical Engineering and Automation of Aalto University. 
	His research interests are in multi-sensor data processing systems with applications in location sensing, health and medical technology, machine learning, inverse problems, and brain imaging. 
	He is a Senior Member of IEEE and serving as an Associate Editor of IEEE Signal Processing Letters.
\end{IEEEbiography}

\end{document}